\documentclass[acmtog]{acmart}

\usepackage{graphicx}
\usepackage{soul}
\usepackage{url}
\usepackage{booktabs}
\usepackage{multirow}
\usepackage{bm}
\usepackage{amssymb}
\usepackage{subcaption}
\usepackage[ruled]{algorithm2e} % For algorithms

\SetAlFnt{\small}
\SetAlCapFnt{\small}
\SetAlCapNameFnt{\small}
\SetAlCapHSkip{0pt}
\IncMargin{-\parindent}

\setcitestyle{square}

\newcommand{\parahead}[1]{\vspace{5pt}\noindent\emph{#1}:\ }
\newcommand\T{\rule{0pt}{2.6ex}}       % Top strut
\newcommand{\ie}{i.e.,\ }

\newcommand{\etal}{~et~al.\ }

\newcommand{\change}[1]{{#1}}

\newcommand{\ignore}[1]{{}}

\newcommand{\FPS}{30}
\newenvironment{packed_itemize}
{\begin{itemize}
    \setlength{\itemsep}{1pt}
    \setlength{\parskip}{0pt}
    \setlength{\parsep}{0pt}
}{\end{itemize}}

\newcommand{\Pprojected}[1]{\mathbf{K}_{#1}} % 2D keypoints
\newcommand{\Pcanonical}[1]{\mathbf{P}^{\text{L}}_{#1}} % 3D root centered
\newcommand{\Ptheta}{\mathbf{P}^{\text{G}}_t(\bm{\theta}, \mathbf{d})}  %Global pose parameters
\newcommand{\Pthetasimple}[1]{\mathbf{P}^{\text{G}}_{#1}}  %Global pose parameters
\newcommand{\Pthetasimpleacc}[1]{\widehat{\mathbf{P}^{\text{G}}_{#1}}} % Acceleration
\newcommand{\Pthetasimplevel}[1]{\widetilde{\mathbf{P}^{\text{G}}_{#1}}} % Acceleration
\newcommand{\skeleton}{S} % skeleton definition, incorporating bone length and angle parametrization.

\begin{document}

%%% This is the ``teaser'' command, which puts an figure, centered, below 
%%% the title and author information, and above the body of the content.
\title{VNect: Real-time 3D Human Pose Estimation with a Single RGB Camera}

% OPTION 1
\author{Dushyant Mehta\textsuperscript{1,2}, Srinath Sridhar\textsuperscript{1}, Oleksandr Sotnychenko\textsuperscript{1},
  Helge Rhodin\textsuperscript{1}, Mohammad Shafiei\textsuperscript{1,2}, Hans-Peter Seidel\textsuperscript{1},
  Weipeng Xu\textsuperscript{1}, Dan Casas\textsuperscript{3}, Christian Theobalt\textsuperscript{1}
}
%\author{asas}
\affiliation{%
  \institution{
    \textsuperscript{1}Max Planck Institute for Informatics, \textsuperscript{2}Saarland University, \textsuperscript{3}Universidad Rey Juan Carlos
  }
}

\renewcommand{\shortauthors}{D. Mehta et. al.}

\begin{abstract}
We present the first real-time method to capture the full global 3D skeletal pose of a human in a stable, temporally consistent \change{manner using} a single RGB camera.
Our method combines a new convolutional neural network (CNN) based pose regressor with kinematic skeleton fitting.
Our \change{novel} fully-convolutional pose formulation regresses 2D and 3D joint positions jointly in real time and does not \change{require} tightly cropped input frames.
\change{A real-time kinematic skeleton fitting method uses the CNN output to }yield temporally stable 3D global pose reconstructions on the basis of a coherent kinematic skeleton.
This makes our approach the first monocular RGB method usable in real-time applications such as 3D character control---thus far, the only monocular methods for such applications employed specialized \mbox{RGB-D} cameras.
Our method's accuracy is quantitatively on par with the best offline \change{3D monocular RGB} pose \change{estimation} methods.
Our results are qualitatively comparable to, and sometimes better than, results from monocular \mbox{RGB-D} approaches, such as the Kinect.
However, we show that our approach is more broadly applicable than \mbox{RGB-D} solutions, \ie \change{it works} for outdoor scenes, \change{community videos, and low quality commodity RGB cameras}.
\end{abstract}

%\begin{CCSXML}
%<ccs2012>
%<concept>
%<concept_id>10010147.10010371.10010352.10010238</concept_id>
%<concept_desc>Computing methodologies~Motion capture</concept_desc>
%<concept_significance>500</concept_significance>
%</concept>
%</ccs2012>
%\end{CCSXML}
%\ccsdesc[500]{Computing methodologies~Motion capture}

% User-generated keywords.
%\keywords{body pose, monocular, real time}
\thanks{This work is was funded by the ERC Starting Grant project CapReal (335545).
Dan Casas was supported by a Marie Curie Individual Fellow, grant agreement 707326.\ \\ \ \\}

% The next three commands are required, and insert the user-generated keywords, 
% The CCS concepts list, and the rights management text.
% Please make sure there is a blank line between each of these three commands.

\begin{teaserfigure}
 \includegraphics[width=\linewidth]{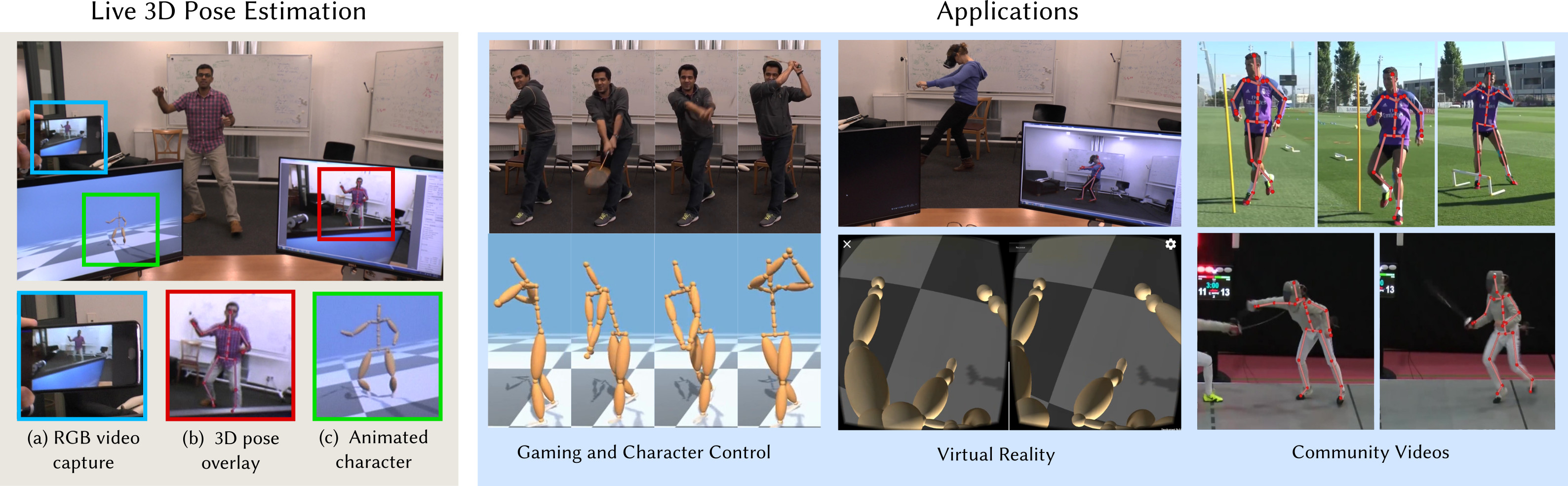}
   \caption{We recover the full global 3D skeleton pose in real-time from a single RGB camera, even wireless capture is possible by streaming from a smartphone (left). It enables applications such as controlling a game character, embodied VR, sport motion analysis and reconstruction of community video (right). Community videos (CC BY) courtesy of Real Madrid C.F.\ \shortcite{ronaldo_recovery_video} and RUSFENCING-TV \shortcite{fencing_video}.}
   \label{fig:teaser}
\end{teaserfigure}

\maketitle

\vspace{-0.3cm}
\section{Introduction}
Optical skeletal motion capture of humans is widely used in applications such as character animation for movies and games, sports and biomechanics, and medicine. To overcome the usability constraints imposed by commercial systems requiring marker suits~\cite{menache2000understanding}, researchers developed marker-less motion capture methods that estimate motion in more general scenes using multi-view video~\cite{moeslund_survey_2006}, \change{with recent solutions being real-time} \cite{stoll_fast_iccv2011}.
The \change{swell in} popularity of applications such as real-time motion-driven 3D game character control, self-immersion in 3D virtual and augmented reality, and human--computer interaction, has led to new real-time full-body motion estimation techniques using only a single, easy to install, depth camera, such as the Microsoft Kinect~\cite{kinectv1,kinectv2,kinectsdk}.
\mbox{RGB-D} cameras provide valuable depth data which greatly simplifies monocular pose reconstruction.
However, \mbox{RGB-D} cameras often fail in general outdoor scenes \change{(due to sunlight interference)}, are bulkier, have higher power consumption, have lower resolution and limited range, and are not as widely and cheaply available as color cameras.

Skeletal pose estimation from a single color camera is a much more challenging and severely underconstrained problem.
\change{Monocular} RGB body pose estimation in 2D has been widely researched, but estimates only the 2D skeletal pose~\cite{wei_cpm_cvpr16,felzenszwalb_pictorial_ijcv05,felzenszwalb_dpm_pami10,bourdev2009poselets,ferrari_pose_cvpr2009,pishchulin_strong_iccv13}.
Learning-based discriminative methods, in particular deep learning methods~\cite{insafutdinov_deepercut_eccv16,newell_stacked_hourglass_eccv16,tompson_cnn_graph_pose_nips14,lifshitz_deep_consensus_eccv16}, represent the current state of the art in 2D pose estimation, with some of these methods demonstrating real-time performance~\cite{wei_cpm_cvpr16,cao2016realtime}.
\change{Monocular} RGB estimation of the 3D skeletal pose is a much harder challenge tackled by \change{relatively fewer} methods~\cite{tekin_motion_comp_cvpr16,tekin_fusion_arxiv16,zhou_deep_kinematic_arxiv16,zhou_convexrelaxation_cvpr2015,zhou_sparseness_deepness_cvpr15,bogo_smpl_eccv16}.
Unfortunately, these methods are typically offline, and they often reconstruct 3D joint positions individually per image, which are temporally unstable, and do not enforce constant bone lengths.
Most approaches also capture local 3D pose relative to a bounding box, and not the full global 3D pose.
\change{This makes them unsuitable} for applications such as real-time 3D character control.

In this paper, we present the first method that captures temporally consistent global {3D} human pose---in terms of joint angles of a single, stable kinematic skeleton---in real-time (30~Hz) from a single RGB video in a general environment.
Our approach builds upon the top performing single RGB 3D pose estimation methods using convolutional neural networks (CNNs)~\cite{pavlakos_volumetric3d_arxiv16,mehta_mlc3d_arxiv16}.
High accuracy requires training comparably deep networks which are harder to run in real-time, \change{partly due to} additional preprocessing steps such as bounding box extraction.
Mehta \etal \shortcite{mehta_mlc3d_arxiv16} use a 100-layer architecture to predict 2D and 3D joint positions simultaneously, but is unsuitable for real-time execution.
To improve runtime, we use a shallower 50-layer network.
However, for best quality at real-time frame rates, we do not merely use a shallower variant, but extend it to a \change{novel} fully-convolutional formulation.
This enables higher accuracy 2D and 3D pose regression, in particular of end effectors (hands, feet), in real-time.
In contrast to existing solutions our approach allows operation on non-cropped images, {and where run-time is a concern, it can be used to bootstrap a simple bounding box tracker.}
We also combine the CNN-based joint position regression with an efficient optimization step to fit a 3D skeleton to these reconstructions in a temporally stable way,{ yielding the global pose and joint angles of the skeleton.}
In summary, we contribute \change{by proposing the first real-time method to capture global 3D kinematic skeleton pose from single RGB video.}
To strike a good compromise between computational complexity and accuracy\change{, our method} combines:
\begin{packed_itemize}
\item A new real-time, fully-convolutional \change{3D body pose formulation using CNNs} that yields 2D and 3D joint positions simultaneously and forgoes the need to perform expensive bounding box computations. 
\item Model-based kinematic skeleton fitting against \change{the} 2D/3D pose predictions to produce temporally stable \change{joint angles of a metric global 3D skeleton}, in real time.
\end{packed_itemize}

Our real-time method achieves state-of-the-art accuracy comparable to the best offline RGB pose estimation methods on standard 3D human body pose benchmarks, particularly for end effector positions (Section~\ref{sec:quantitative}). Our results are qualitatively comparable to, and sometimes better than, state-of-the-art single \mbox{RGB-D} methods~\cite{girshick2011efficient}, even commercial ones~\cite{kinectsdk}. 
We experimentally show that this makes ours the first single-RGB method usable for similar real-time 3D applications---so far only feasible with \mbox{RGB-D} input---such as game character control or immersive first person virtual reality (VR). We further show that our method succeeds \change{in settings where existing \mbox{RGB-D} methods would not, such as} outdoor scenes, community videos, and even with \change{low quality video streams from} ubiquitous mobile phone cameras. 

\begin{figure*}[t]
  \includegraphics[width=\linewidth]{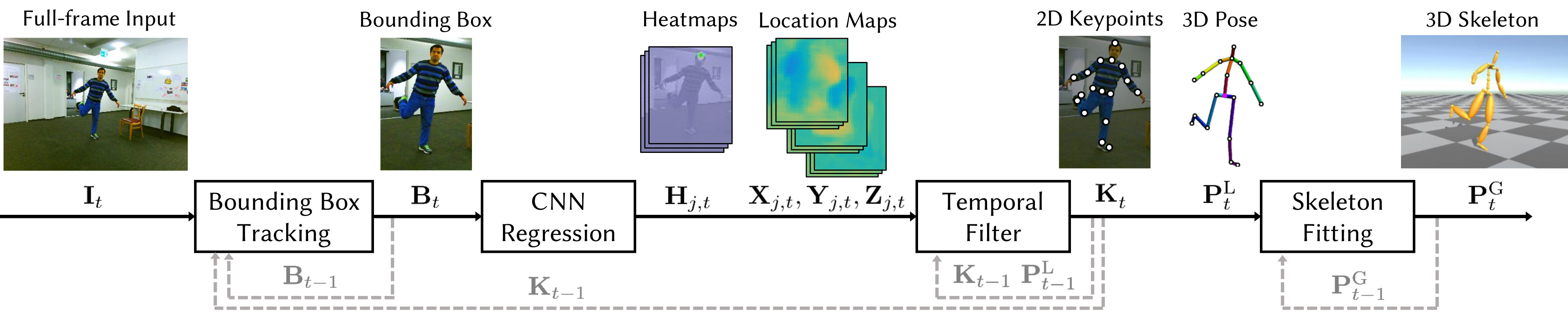}
\caption{Overview. Given a full-size image $\mathbf{I}_t$ at frame $t$, the person-centered crop $\mathbf{B}_t$ is efficiently extracted by bounding box tracking, using the previous frame's keypoints $\mathbf{K}_{t-1}$. From the crop, the CNN jointly predicts 2D heatmaps $\mathbf{H}_{j,t}$ and our novel 3D \textit{location-maps} $\mathbf{X}_{j,t},\mathbf{Y}_{j,t}$ and $\mathbf{Z}_{j,t}$ for all joints $j$.
The 2D keypoints $\mathbf{K}_t$ are retrieved from $\mathbf{H}_{j,t}$ and, after filtering, are used to read off 3D pose $\mathbf{P}^\text{L}_t$  from $\mathbf{X}_{j,t},\mathbf{Y}_{j,t}$ and $\mathbf{Z}_{j,t}$.
These per-frame estimates are combined to stable global pose $\mathbf{P}^\text{G}_t$ by skeleton fitting. Information from frame $t-1$ is marked in gray-dashed.}
   \label{fig:overview}
\end{figure*}

%%%%%%%%%%%%%%%%%%%%%%%%%%%%%%%%%%%%%%%%%%%%%%%%%%%%%%%%%%%%%%%%%%%%%%%%%%%%%%%%%%%%
\section{Related Work}
%%%%%%%%%%%%%%%%%%%%%%%%%%%%%%%%%%%%%%%%%%%%%%%%%%%%%%%%%%%%%%%%%%%%%%%%%%%%%%%%%%%%
Our goal is stable 3D skeletal motion capture from (1) a single camera (2) in real-time. We focus the discussion of related work on approaches from the large body of marker-less motion capture research that contributed to attaining either of these properties.

\parahead{Multi-view}
With multi-view setups markerless motion-captue solutions attain high accuracy.
Tracking of a manually initialized actor model from frame to frame with a generative image formation \change{model} is common. 
See \cite{moeslund_survey_2006} for a complete overview.
Most methods target high quality with offline computation   \cite{bregler1998tracking,howe1999bayesian,sidenbladh2000stochastic,starck_model_iccv2003,loper2014opendr}.
Real-time performance can be attained by representing the actor with Gaussians \cite{wren_pfinder_pami1997,stoll_fast_iccv2011,rhodin_versatile_iccv2015} and other approximations \cite{ma2014realtime}, in addition to formulations allowing model-to-image fitting.
However, these tracking-based approaches often lose track in local minima of the non-convex fitting functions they optimize
and require separate initialization, e.g. using \cite{sminchisescu_covariance_cvpr2001,bogo_smpl_eccv16,rhodin_general_eccv16}.
Robustness could be increased with a combination of generative and discriminative estimation \cite{elhayek_convmocap_TPAMI2016}, even from a single input view \cite{sminchisescu_learning_cvpr2006,rosales2006combining}, and egocentric perspective \cite{rhodin_egocap_SIGGRAPHAsia2016}.
We utilize generative tracking components to ensure temporal stability, but avoid model projection through a full image formation model to speed up estimation.
Instead, we combine discriminative pose estimation with kinematic fitting to succeed in our underconstrained setting.

\parahead{Monocular Depth-based}
The additional depth channel provided by \mbox{RGB-D} sensors has led to robust real-time pose estimation solutions \cite{baak_posedepth_iccv11,ganapathi2012real,wei_single-rgbd_tog12,shotton2013real,ma2014realtime,ye_single-rgbd_cvpr14} and the availability of low-cost devices has enabled a range of new applications. %\HR{Srinath, can you add 2-3 important hand depth-based papers.}
Even real-time tracking of general deforming objects \cite{zollhoefer2014deformable} and template-free reconstruction \cite{newcombe2015dynamic,innmann2016volume,orts2016holoportation,dou2016fusion4d} has been demonstrated.
\mbox{RGB-D} information overcomes forward-backwards ambiguities in monocular pose estimation.
Our goal is a video solution that overcomes depth ambiguities without relying on a specialized active sensor.

\parahead{Monocular RGB}
Monocular generative motion capture has only been shown for short clips and when paired with strong motion priors \cite{urtasun_temporal_cviu2006} or in combination with discriminative re-initialization \cite{sminchisescu_learning_cvpr2006,rosales2006combining}, since generative reconstruction is fundamentally underconstrained.
Using photo-realistic template models for model fitting enables more robust monocular tracking of simple motions, but requires more expensive offline computation \cite{de2008model}.
Sampling-based methods avoid local minima \cite{deutscher2005articulated,balan2005quantitative,gall_optimization_ijcv2010,bo2010twin}. However, real-time variants can not guarantee global convergence due to a limited number of samples, such as particle swarm optimization techniques \cite{oikonomidis2011efficient}.
Structure-from-motion techniques exploit motion cues in a batch of frames \cite{garg2013dense}, and have also been applied to human motion estimation \cite{gotardo2011computing,lee2013procrustean,park20113d,zhu20113d}.
However, batch optimization does not apply to our real-time setting, where frames are streamed sequentially.
For some applications manual annotation and correction of frames is suitable, for instance to enable movie actor reshaping \cite{jain2010movie} and garment replacement in video \cite{rogge2014garment}. In combination with physical constraints, highly accurate reconstructions are possible from monocular video \cite{wei2010videomocap}. 
Vondrak \etal \shortcite{vondrak2012video} succeed without manual annotation by simulating biped-controllers, but require batch-optimization.
While these methods can yield high-quality reconstructions, interaction and expensive optimization preclude live applications.

Discriminative 2D human pose estimation is often an intermediate step to monocular 3D pose estimation. Pictorial structure approaches infer body part locations by
message passing over a huge set of pose-states
\cite{agarwal_recovering_pami06,felzenszwalb_pictorial_ijcv05,ferrari_pose_cvpr2009,andriluka_pictorial_cvpr09,bourdev2009poselets,johnson_lsp_bmvc10} and have been extended to 3D pose estimation \cite{balan_detailed_cvpr2007,sigal_loose_ijcv2012,amin_multi_bmvc2013,belagiannis_3d_cvpr2014}.
Recent approaches outperform these methods in computation time and accuracy by leveraging large image databases with 2D joint location annotation, which enables high accuracy prediction with deep CNNs \cite{hu_bottomup_cvpr16,belagiannis_recurrent_arxiv6,pishchulin_deepcut_cvpr16,insafutdinov_deepercut_eccv16,wei_cpm_cvpr16},
on multiple GPUs, even at real-time rates \cite{cao2016realtime}. 
\change{Given 2D joint locations, lifting them to 3D pose is challenging}. Existing approaches use bone length and depth ordering constraints \cite{taylor_articulated_cvpr00,mori_contexts_pami06},
sparsity assumptions \cite{wang_robust_cvpr2014,zhou_convexrelaxation_cvpr2015,zhou_sparse_arXiv2015},
joint limits \cite{akhter_pose_conditioned_cvpr15},
inter-penetration constraints \cite{bogo_smpl_eccv16},
temporal dependencies \cite{rhodin_general_eccv16},
and
regression \cite{yasin_dual_source_cvpr16}.
Treating 3D pose as a hidden variable in 2D estimation is an alternative \cite{brau_annotations_3dv16}.
However, the sparse set of 2D locations loses image evidence, e.g. on forward-backwards orientation of limbs, which leads to erroneous estimates in ambiguous cases.
To overcome these ambiguities, discriminative methods have been proposed that learn implicit depth features for
3D pose directly from more expressive image representations.
Rosales and Sclaroff regress 3D pose from silhouette images with the \emph{specialized
mappings architecture} \shortcite{rosales2000specialized},
Agarwal and Triggs with linear regression \shortcite{agarwal_recovering_pami06},
and Elgammal and Lee through a joint embedding of images and 3D pose \shortcite{elgammal2004inferring}.
Sminchisescu further utilized temporal consistency to propagate pose probabilities with a \emph{Bayesian mixture of experts Markov model} \shortcite{sminchisescu2007bm3e}.
\change{Relying on the recent advances in machine learning techniques and compute capabilities, approaches for direct 3D pose regression from the input image have been proposed, 
using structured learning of latent pose}
\cite{tekin_structured_bmvc16,li_maxmargin_iccv15}, joint prediction of 2D and 3D pose~\cite{li_accv14,tekin_fusion_arxiv16,yasin_dual_source_cvpr16},
transfer of features from 2D datasets \cite{mehta_mlc3d_arxiv16}, 
novel pose space formulations \cite{pavlakos_volumetric3d_arxiv16}
and classification over example poses \cite{pons_posebits_cvpr14,rogez_mocap_arxiv16}. 
Relative per-bone predictions \cite{li_accv14}, 
kinematic skeleton models \cite{zhou_deep_kinematic_arxiv16},
or root centered joint positions \cite{ionescu_iterated_cvpr14} are used as the eventual output space. 
Such direct 3D pose regression methods capture depth relations well, but 3D estimates usually do not accurately match the true 2D location when re-projected to the image, because estimations are done in cropped images that lose camera perspective effects, using a canonical height, and minimize 3D loss instead of projection to 2D.
Furthermore, they only deliver joint positions, are temporally unstable, and none has shown real-time performance.
We propose a method to combine 2D and 3D estimates in real-time along with temporal tracking.
It is inspired by the method of Tekin et al.~\shortcite{tekin_motion_comp_cvpr16}, where batches of frames are processed offline after motion compensation, and is related to the recently proposed per-frame combination of 2D and 3D pose \cite{tekin_fusion_arxiv16}.

Notably, only few methods target real-time monocular reconstruction. Exceptions are the 
 regression of 3D pose from Haar features by Bissacco et al.~\shortcite{bissacco2007fast} and
  detection of a set of discrete poses from edge direction histograms in the vicinity of the previous frame pose \cite{taycher2006conditional}.
 Both only obtain temporally unstable, coarse pose, not directly usable in our applications.
Chai and Hodgins obtain sufficient quality to drive virtual avatars in real-time, but require visual markers \cite{chai2005performance}.
The use of CNNs in real time has been explored for
variants of the object detection problem, for instance bounding box detection and pedestrian detection methods have leveraged application specific architectures~\cite{wei_ssd_eccv16,redmon_yolo_arxiv15,angelova_realtime_pedestrian_bmvc15} and preprocessing steps~\cite{ren_faster_rcnn_nips15}.

In a similar vein, we propose a 3D pose estimation approach that leverages a novel fully-convolutional CNN formulation to predict 2D and 3D pose jointly. In combination with inexpensive preprocessing and an optimization based skeletal fitting method, it enables high accuracy pose estimation, while running at more than \FPS~Hz.

\section{Overview}
Our system is capable of obtaining a temporally consistent, full 3D skeletal pose of a human from a monocular RGB camera.
Estimating 3D pose from a single RGB camera is a challenging, underconstrained problem with inherent ambiguities. 
Figure~\ref{fig:overview} provides an overview of our method to tackle this challenging problem.
It consists of two primary components.
The first is a convolutional neural network (CNN) to regress 2D and 3D joint positions under the ill-posed monocular capture conditions. It is trained on annotated 3D human pose datasets~\cite{ionescu_human36_pami14,mehta_mlc3d_arxiv16}, additionally leveraging annotated 2D human pose datasets \cite{andriluka_mpii2d_cvpr14,johnson_lsp_bmvc10} for improved in-the-wild performance. 
The second component combines the regressed joint positions with a kinematic skeleton fitting method to produce a temporally stable, camera-relative, full 3D skeletal pose.

\parahead{CNN Pose Regression}
The core of our method is a CNN that predicts both 2D, and root (pelvis) relative 3D joint positions in real-time.
The new proposed fully-convolutional pose formulation leads to results on par with the state-of-the-art offline methods in 3D joint position accuracy (see Section~\ref{sec:quantitative} for details). Being fully-convolutional, it can operate in the absence of tight crops around the subject. 
The CNN is capable of predicting joint positions for a diverse class of activities regardless of the scene settings, providing a strong basis for further pose refinement to produce temporally consistent full-3D pose parameters 

\parahead{Kinematic Skeleton Fitting}
The 2D and the 3D predictions from the CNN, together with the temporal history of the sequence, can be leveraged to obtain temporally consistent full-3D skeletal pose, with the skeletal root (pelvis) localized in camera space.  
Our approach uses an optimization function that: (1) combines the predicted 2D and 3D joint positions to fit a kinematic skeleton in a least squares sense, (2) ensures temporally smooth tracking over time. 
We further improve the stability of the tracked pose by applying filtering steps at different stages.

\parahead{Skeleton Initialization (Optional)}
The system is set up with a default skeleton which works well out of the box for most humans.
For more accurate estimates, the relative body proportions of the underlying skeleton \change{can be} adapted to that of the subject, by averaging CNN predictions for a few frames at the beginning.
Since monocular reconstruction is ambiguous without a scale reference,
the CNN predicts height normalized 3D joint positions.
Users only need to provide their height (distance from head to toe) once, so that we can track the 3D pose in true metric space.

\section{Real-time Monocular 3D Pose Estimation}
In this section, we describe in detail the different components of our method to estimate a temporally consistent 3D skeletal motion from monocular RGB input sequences.
As input, we assume a continuous stream of monocular RGB images $\{...,\mathbf{I}_{t-1},\mathbf{I}_{t}\}$.
For frame $t$ in the input stream, the final output of our approach is $\Pthetasimple{t}$ which is the full global 3D skeletal pose of the person being tracked.
Because this output is already temporally consistent and in global 3D space, it can be readily used in applications such as character control.

We use the following notation for the output in the intermediate components of our method.
The CNN pose regressor jointly estimates the 2D joint positions $\Pprojected{t}$ and root-relative 3D joint positions ${\Pcanonical{t}}$ (Section \ref{sec:network}).
The 3D skeleton fitting component combines the 2D and 3D joint position predictions to estimate a smooth, temporally consistent pose $\Ptheta$, which is parameterized by the global position $\mathbf{d}$ in camera space, and joint angles $\bm{\theta}$ of the kinematic skeleton $\skeleton$. $J$ indicates the number of joints. We drop the frame-number subscript $t$ in certain sections to aid readability.

\subsection{CNN Pose Regression}
\label{sec:network}	

The goal of CNN pose regression is to obtain joint positions, both, in 2D image space and 3D.
For 2D pose estimation with neural nets, the change of formulation from direct regression of $x,y$ body-joint coordinates \cite{toshev_deeppose_cvpr14} to a heatmap based body-joint detection formulation \cite{tompson_cnn_graph_pose_nips14} has been the key driver behind the recent developments in 2D pose estimation. The heatmap based formulation naturally ties image evidence to pose estimation by predicting a confidence heatmap $\mathbf{H}_{j,t}$ over the image plane for each joint $j \in \{1 .. J\}$.

Existing approaches to 3D pose estimation lack such an image-to-prediction association, often directly regressing the root-relative joint locations \cite{ionescu_iterated_cvpr14}, leading to predicted poses whose extent of articulation doesn't reflect that of the person in the image. See Figure \ref{fig:comparison}. Treating pose as a vector of joint locations also causes a natural gravitation towards networks with fully-connected formulations \cite{tekin_structured_bmvc16,mehta_mlc3d_arxiv16,rogez_mocap_arxiv16,yu_mono_heightmap_eccv16}, restricting the inputs to tight crops at a fixed resolution, a limitation that needs to be overcome. These methods assume the availability of tight bounding boxes, which necessitates supplementation with separate bounding box estimators for actual usage, which further adds to the run-time of these methods. The fully-convolutional formulation of Pavlakos \etal \cite{pavlakos_volumetric3d_arxiv16} seeks to alleviate some of these issues, but is limited by the expensive per-joint volumetric formulation, which \change{still relies on cropped input} and does not scale well to larger image sizes.%, and high the need for fixed size input.

\begin{figure}[]
  %% \centering\includegraphics*[page=1,width=1\linewidth]{Figures/method.png}
  	\centering\includegraphics*[page=1,width=1\linewidth]{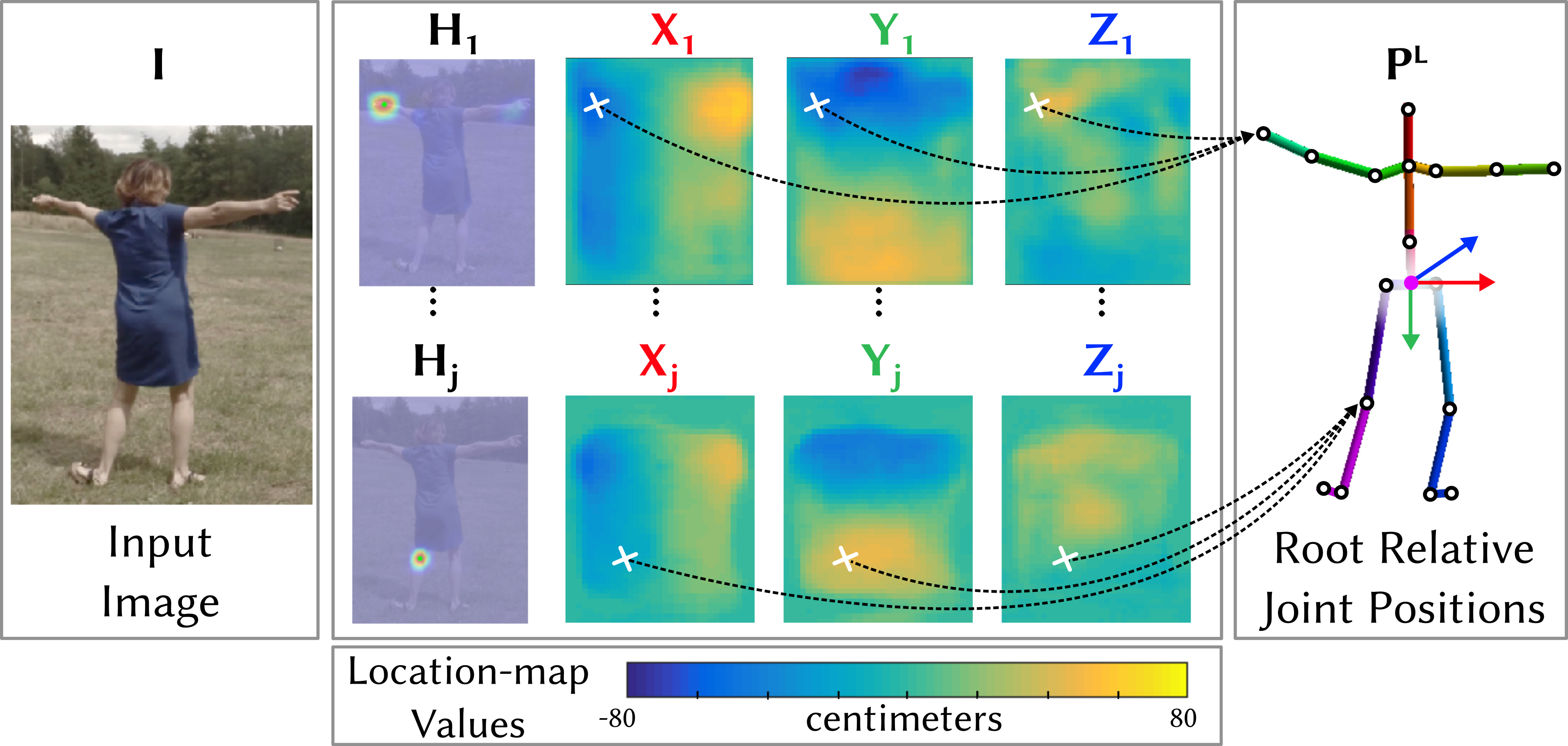}
	\caption{Schema of the fully-convolutional formulation for predicting root relative joint locations. For each joint $j$, the 3D coordinates are predicted from their respective \textit{location-maps} $\mathbf{X}_j, \mathbf{Y}_j, \mathbf{Z}_j$ at the position of the maximum in the corresponding 2D heatmap $\mathbf{H}_j$. The structure observed here in the location-maps emerges due to the spatial loss formulation. See Section \ref{sec:network}.}
	\label{fig:pose_formulation}
\end{figure}

We overcome these limitations through our new formulation, by extending the 2D heatmap formulation to 3D using three additional \textit{location-maps} $\mathbf{X}_j, \mathbf{Y}_j, \mathbf{Z}_j$ per joint $j$, capturing the root-relative locations $x_j$, $y_j$ and $z_j$  respectively. %\HR{link to $P^C$ is missing}
To have the 3D pose prediction linked more strongly to the 2D appearance in the image, the $x_j$, $y_j$ and $z_j$ values are read off from their respective location-maps at the position of the maximum of the corresponding joint's 2D heatmap $\mathbf{H}_j$, and stored in ${\Pcanonical{}}$ = \{$\mathbf{x}$, $\mathbf{y}$, $\mathbf{z}$\}, where $\mathbf{x} \in \mathbb{R}^{1 \times J}$ is a vector that stores the coordinate $x$ location of each joint maximum.
The pose formulation is visualized in Figure \ref{fig:pose_formulation}. Networks using this fully-convolutional formulation are not constrained in input image size, and can work without tight crops. Additionally, the network provides 2D and 3D joint location estimates without additional overhead, which we exploit in \change{subsequent steps} for real-time estimation.
Section \ref{sec:quantitative} shows the improvements afforded by this formulation.
%\

\parahead{Loss Term}
To enforce the fact that we are only interested in $x_j$, $y_j$ and $z_j$ from their respective maps at joint $j$'s 2D location, the joint location-map loss is weighted stronger around the joint's 2D location. 
We use the L2 loss. For $x_j$ is the loss formulation is
\begin{equation}
\label{eq:loss}
\text{Loss}(x_j) = \|\mathbf{H}^\text{GT}_j \odot (\mathbf{X}_j - \mathbf{X}^\text{GT}_j)\|_2
\text{,}
\end{equation}
where $\text{GT}$ indicates ground truth and $\odot$ is the Hadamard product.
The location maps are weighted with the respective ground truth 2D heatmap $\mathbf{H}^\text{GT}_j$, which in turn
%are constructed to 
have confidence equal to a Gaussian with a small support localized at joint $j$'s 2D location.
Note that no structure is imposed on the location-maps. The structure that emerges in the predicted location-maps is indicative of the correlation of $x_j$ and $y_j$ with root relative location of joint $j$ in the image plane. 
See Figure \ref{fig:pose_formulation}.

\begin{figure*}[t]
  \includegraphics[width=0.95\linewidth]{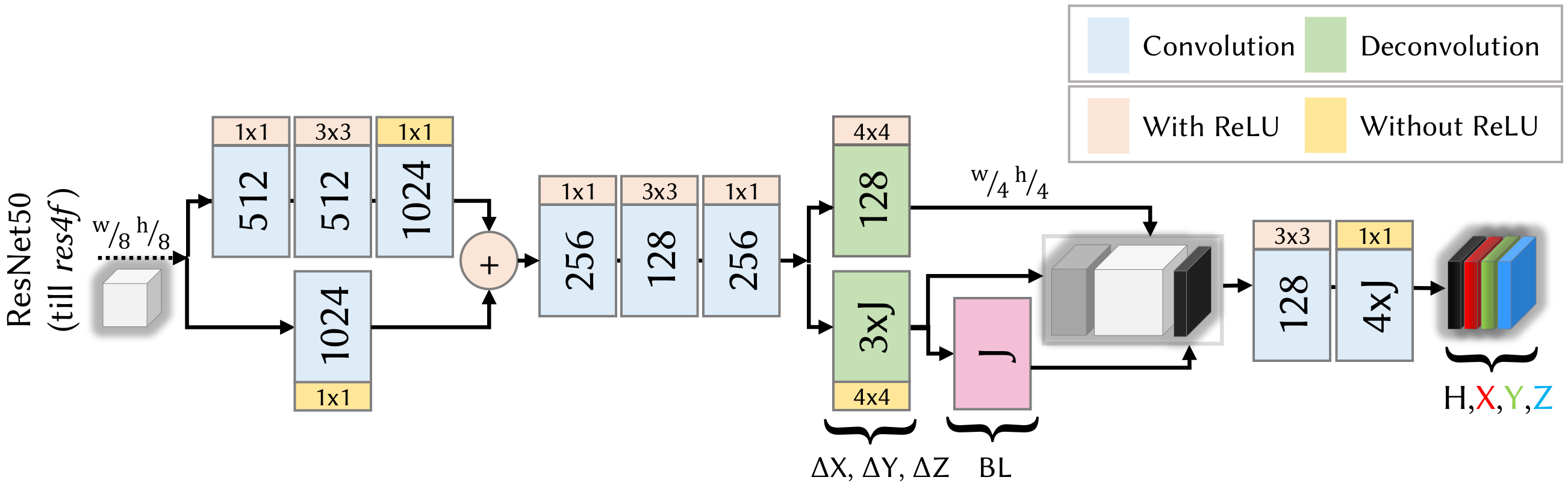}
\caption{Network Structure. The structure above is preceded by ResNet50/100 till level 4. We use kinematic parent relative 3D joint location predictions $\Delta\mathbf{X}$, $\Delta\mathbf{Y}$, $\Delta\mathbf{Z}$ as well as bone length maps $\mathbf{BL}$ constructed from these as auxiliary tasks.The network predicts 2D location heatmaps $\mathbf{H}$ and root relative 3D joint locations $\mathbf{X}$, $\mathbf{Y}$, $\mathbf{Z}$. Refer to Section \ref{sec:network}.}
   \label{fig:network_structure}
\end{figure*}

\parahead{Network Details}
We use the proposed formulation to adapt the ResNet50 network architecture of He \etal \shortcite{he_resnet_cvpr2016}. We replace the layers of ResNet50 from \emph{res5a} onwards with the architecture depicted in Figure \ref{fig:network_structure}, producing the heatmaps and location-maps for all joints $j \in \{1..J\}$. \change{After training, the Batch Normalization~\cite{ioffe_batchnorm_icml15} layers are merged with the weights of their preceding convolution layers to improve the speed of the forward pass.}

\parahead{Intermediate Supervision}
We predict the 2D heatmaps and 3D location-maps from the features at \textit{res4d} and \textit{res5a}, tapering down the weights of intermediate losses with increasing iteration count. 
Additionally, similar to the root-relative location-maps $\mathbf{X}_j$, $\mathbf{Y}_j$ and $\mathbf{Z}_j$, we predict kinematic parent-relative location-maps $\mathbf{\Delta X}_j$, $\mathbf{\Delta Y}_j$ and $\mathbf{\Delta Z}_j$ from the features at \emph{res5b} and compute bone length-maps as:
\begin{equation}
\label{eq:bonemap}
\mathbf{BL}_j = \sqrt[]{\mathbf{\Delta X}_j \odot \mathbf{\Delta X}_j + \mathbf{\Delta Y}_j \odot \mathbf{\Delta Y}_j + \mathbf{\Delta Z}_j \odot \mathbf{\Delta Z}_j }
\text{.}
\end{equation}
\begin{figure}[t!]
	\centering\includegraphics*[page=1,width=1.0\linewidth]{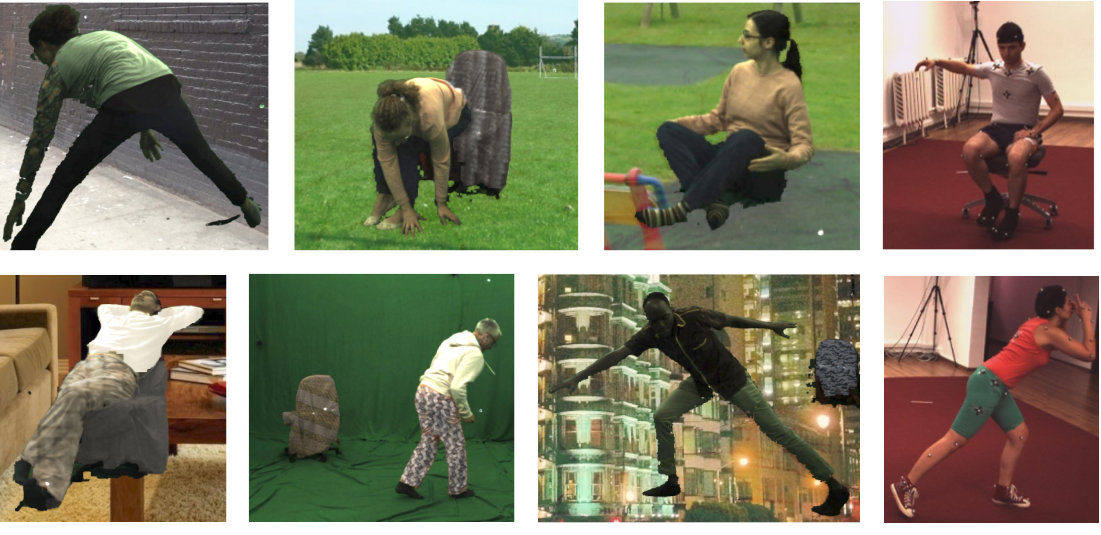}
	\caption{Representative training frames from Human3.6m and MPI-INF-3DHP 3D pose datasets. Also shown are the background, clothing and occluder augmentations done on MPI-INF-3DHP training data.}
	\label{fig:augmentation}
\end{figure}
These intermediate predictions are subsequently concatenated with the intermediate features, to give the network an explicit notion of bone lengths to guide the predictions. See Figure \ref{fig:network_structure}.

Experiments showed that the deeper variants of ResNet offer only small gains for a substantial increase (1.5$\times$) in computation time, prompting us to choose ResNet50 to enable real-time, yet highly accurate joint location estimation with the proposed formulation.

\parahead{Training}
The network is pretrained for 2D pose estimation on MPII \cite{andriluka_mpii2d_cvpr14} and LSP \cite{johnson_lsp_bmvc10,johnson_lspet_cvpr11} to allow superior in-the-wild performance, as proposed by Mehta \etal \cite{mehta_mlc3d_arxiv16}. For 3D pose, we use MPI-INF-3DHP~\cite{mehta_mlc3d_arxiv16} and Human3.6m~\cite{ionescu_human36_pami14}. We take training sequences for all subjects except S9 and S11 from Human3.6m. We sample frames as per \cite{ionescu_iterated_cvpr14}. For MPI-INF-3DHP, we consider all 8 training subjects. We choose sequences from all 5 chest-high cameras, 2 head-high cameras (angled down) and 1 knee-high camera (angled up) to learn some degree of invariance to the camera viewpoint. The sampled frames have at least one joint move by $> 200mm$ between them. We use various combinations of background, occluder (chair), upper-body clothing and lower-body clothing augmentation for ~70\% of the selected frames. We train with person centered crops, and use image scale augmentation at 2 scales ($0.7\times , 1.0\times$), resulting in 75k training samples for Human3.6m and 100k training samples for MPI-INF-3DHP. Figure \ref{fig:augmentation} shows a few representative frames of training data.
%.
In addition to the 17 joints typically considered, we use foot tip positions. The ground truth joint positions are with respect to a height normalized skeleton (knee--neck height 92~cm). We make use of the Caffe~\shortcite{jia_caffe_ICM} framework for training, and use the Adadelta~\cite{zeiler2012adadelta} solver with learning rate tapered down with increasing iterations. 
%\

\parahead{Bounding Box Tracker}
Existing offline solutions process each frame in a separate person-localization and bounding box (BB) cropping step \cite{tekin_motion_comp_cvpr16,mehta_mlc3d_arxiv16} or assume bounding boxes are available \cite{tekin_structured_bmvc16,li_maximum_iccv2015,li_accv14,pavlakos_volumetric3d_arxiv16,zhou_deep_kinematic_arxiv16}.
Although our fully-convolutional formulation allows the CNN to work without requiring cropping, the run-time of the CNN is highly dependent on the input image size. Additionally, the CNN is trained for subject sizes in the range of 250--340~px in the frame, requiring averaging of predictions at multiple image scales per frame (scale space search) if processing the full frame at each time step. 
Guaranteeing real-time rates necessitates restricting the size of the input to the network and tracking the scale of the person in the image to avoid searching the scale space in each frame. 
We do this in an integrated way.
The 2D pose predictions from the CNN at each frame are used to determine the BB for the next frame through a slightly larger box around the predictions. 
\change{The smallest rectangle containing the keypoints $\mathbf{K}$ is computed and augmented with a buffer area $0.2\times$ the height vertically and $0.4\times$ the width horizontally. To stabilize the estimates, the BB is shifted horizontally to the centroid of the 2D predictions,}
 and its corners are filtered with a weighted average with the previous frame's BB using a momentum of $0.75$. 
To normalize scale, the BB crop is resized to 368x368 px.
The BB tracker starts with (slow) multi-scale predictions on the full image for the first few frames, and hones in on the person in the image making use of the BB-agnostic predictions from the fully convolutional network. 
The BB tracking is easy to implement and without runtime overhead, since the proposed fully-convolutional network outputs 2D and 3D pose jointly and operates on arbitrary input sizes.

\subsection{Kinematic Skeleton Fitting}
\label{sec:tracker}

Applying per-frame pose estimation techniques on a video does not exploit and ensure temporal consistency of motion, and small pose inaccuracies lead to temporal jitter, an unacceptable artifact for most graphics applications.
We combine the 2D and 3D joint positions in a joint optimization framework, along with temporal filtering and smoothing, to obtain  
an accurate, temporally stable and robust result.
First, the 2D predictions $\Pprojected{t}$ are temporally filtered \cite{casiez_1euro_sigchi12} and used to obtain the 3D coordinates of each joint from the location-map predictions, giving us $\Pcanonical{t}$. To ensure skeletal stability, the bone lengths inherent to $\Pcanonical{t}$ are replaced by the bone lengths of the underlying skeleton in a simple retargeting step that preserves the bone directions of $\Pcanonical{t}$. The resulting 2D and 3D predictions are combined by minimizing the objective energy
\begin{align}
  E_{\text{total}}(\bm{\theta}, \mathbf{d}) &= E_{\text{IK}}(\bm{\theta}, \mathbf{d}) + E_{\text{proj}}(\bm{\theta}, \mathbf{d})\nonumber\\
  &+ E_{\text{smooth}}(\bm{\theta}, \mathbf{d}) + E_{\text{depth}}(\bm{\theta}, \mathbf{d}),
\end{align}
for skeletal joint angles $\bm{\theta}$ and the root joint's location in camera space $\mathbf{d}$. 
The 3D inverse kinematics term $E_{\text{IK}}$ determines the overall pose by similarity to the 3D CNN output $\Pcanonical{t}$.
The projection term $E_{\text{proj}}$ determines global position $\mathbf{d}$ and corrects the 3D pose by re-projection onto the detected 2D keypoits $\mathbf{K_t}$. 
Both terms are implemented with the L2 loss,
\begin{equation}
E_{\text{proj}} = \| \Pi (\Pthetasimple{t}) - \Pprojected{t} \|_2 \text{ and }
E_{\text{IK}} = \|(\Pthetasimple{t} - \mathbf{d}) -\Pcanonical{t} \|_2
\text{,}
\end{equation}
where $\Pi$ is the projection function from 3D to the image plane, and $\Pthetasimple{t} = \Ptheta$. We assume the pinhole projection model. If the camera calibration is unknown a vertical field of view of 54 degrees is assumed. 
Temporal stability is enforced with smoothness prior $E_{\text{smooth}} =  \| \Pthetasimpleacc{t}\|_2$, penalizing the acceleration $\Pthetasimpleacc{t}$. To counteract the strong depth uncertainty in monocular reconstruction, we penalize large variations in depth additionally with
$E_{\text{depth}} =  
%\| \dot{[\Pthetasimple{t}]_z} \|_2$, 
\|[\Pthetasimplevel{t}]_z \|_2$
where 
%$[\dot{\Pthetasimple{t}}]_z$ 
$[\Pthetasimplevel{t}]_z$
is the $z$ component of 3D velocity 
%$\dot{\Pthetasimple{t}}$.
$\Pthetasimplevel{t}$.
Finally, the 3D pose is also filtered with the 1 Euro filter~\cite{casiez_1euro_sigchi12}.
 
\parahead{Parameters}
The energy terms $E_{\text{IK}}, E_{\text{proj}}, E_{\text{smooth}}$ and $E_{\text{depth}}$ are weighted with $\omega_{\text{IK}}=1, \omega_{\text{proj}}=44, \omega_{\text{smooth}}=0.07$ and $\omega_{\text{depth}}=0.11$, respectively.
The parameters of the 1 Euro Filter~\cite{casiez_1euro_sigchi12} are empirically set to $f_{c_\text{min}}=1.7$, $\beta=0.3$ for filtering $\mathbf{K}_t$, 
to $f_{c_\text{min}}=0.8$, $\beta=0.4$ for $\Pcanonical{t}$, 
and to $f_{c_\text{min}}=20$, $\beta=0.4$ for filtering $\Pthetasimple{t}$.
\change{Our implementation uses the Levenberg-Marquardt algorithm from the 	Ceres library~\cite{ceres-solver}.}

\section{Results}
\label{sec:results}
We show live applications of our system at \FPS~Hz.
The reconstruction quality is high \change{and we demonstrate the usefulness of our method 3D character control, embodied virtual reality, and pose tracking from low quality smartphone camera streams}.
See Section \ref{sec:applications} and Figure \ref{fig:teaser}.
Results are best observed in motion in the supplemental video.
The importance of the steps towards enabling these applications with a video solution
are thoroughly \change{evaluated in more than 10 sequences.}
Results are \change{comparable} in quality to depth-camera based \change{solutions like} the Kinect~\cite{kinectv2}
and significantly outperform existing monocular video-based solutions. 
As the qualitative baseline we choose the state-of-the-art 2D to 3D lifting approach of Zhou~\etal~\shortcite{zhou_convexrelaxation_cvpr2015} and the 3D regression approach of Mehta~\etal~\shortcite{mehta_mlc3d_arxiv16}, which estimate joint-positions offline.
The accuracy improvements are further quantitatively validated on the established H3.6M dataset~\cite{ionescu_human36_pami14} and the MPI-INF-3DHP dataset~\cite{mehta_mlc3d_arxiv16}.
The robustness to diverse persons, clothing and scenes is demonstrated on several \change{real-time examples and community} videos.
Please see our project webpage for more results and details\footnote{\normalsize \url{http://gvv.mpi-inf.mpg.de/projects/VNect/}}.

\change{Computations are performed on a 6-core Xeon CPU, 3.8 GHz and a single Titan X (Pascal architecture) GPU}.
The CNN computation takes $\approx$18~ms, the skeleton fitting $\approx$7--10~ms, and preprocessing and filtering 5~ms.

\subsection{Comparison with Active Depth Sensors (Kinect)}
\begin{figure}
\center
\includegraphics[width=0.192\columnwidth,trim=1cm 0cm 4.5cm 0cm,clip]{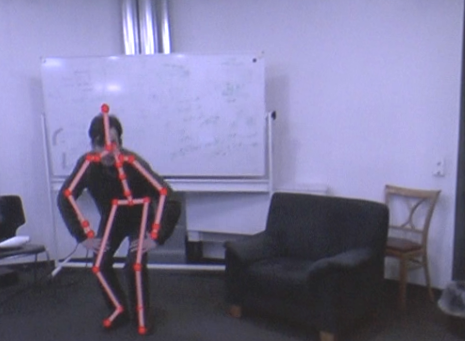}
\includegraphics[width=0.192\columnwidth,trim=1cm 0cm 4.5cm 0cm,clip]{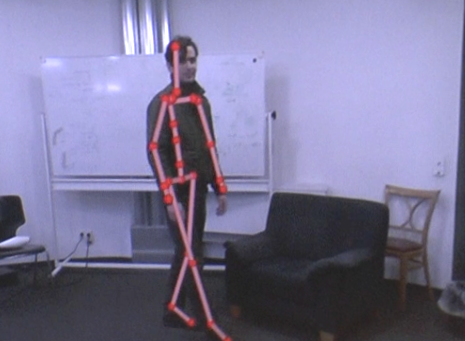}
\includegraphics[width=0.192\columnwidth,trim=1cm 0cm 4.5cm 0cm,clip]{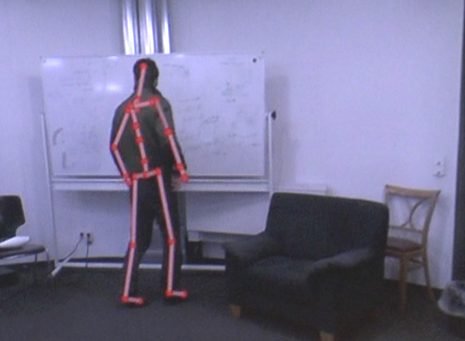}
\includegraphics[width=0.192\columnwidth,trim=1cm 0cm 4.5cm 0cm,clip]{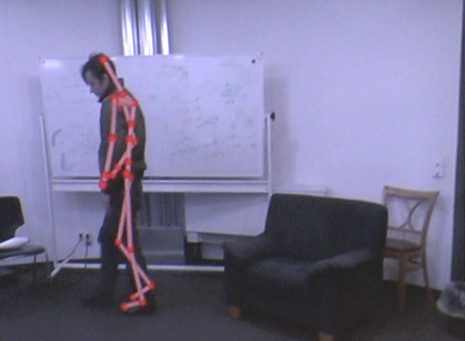}
\includegraphics[width=0.192\columnwidth,trim=1cm 0cm 4.5cm 0cm,clip]{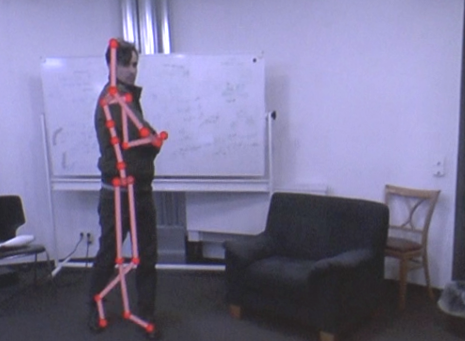}

\vspace{0.05cm}
\includegraphics[width=0.192\columnwidth,trim=1cm 0cm 4.5cm 0cm,clip]{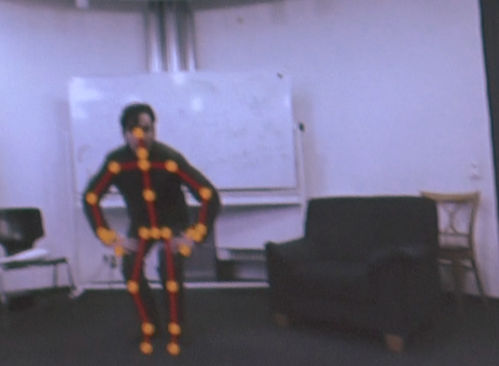}
\includegraphics[width=0.192\columnwidth,trim=1cm 0cm 4.5cm 0cm,clip]{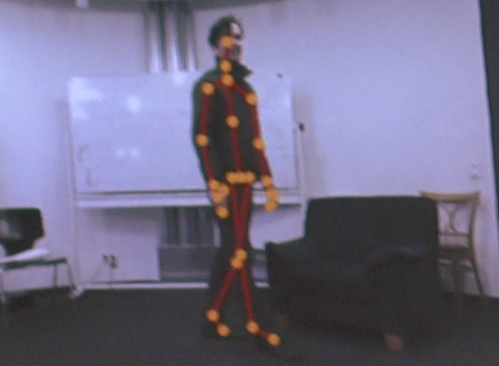}
\includegraphics[width=0.192\columnwidth,trim=1cm 0cm 4.5cm 0cm,clip]{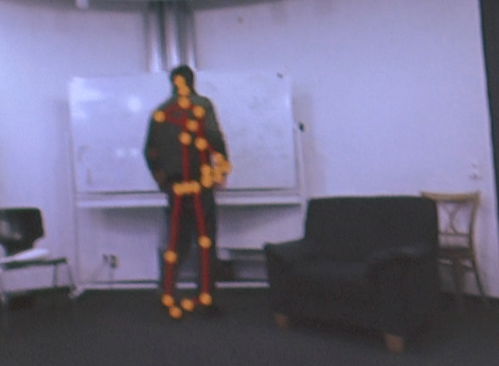}
\includegraphics[width=0.192\columnwidth,trim=1cm 0cm 4.5cm 0cm,clip]{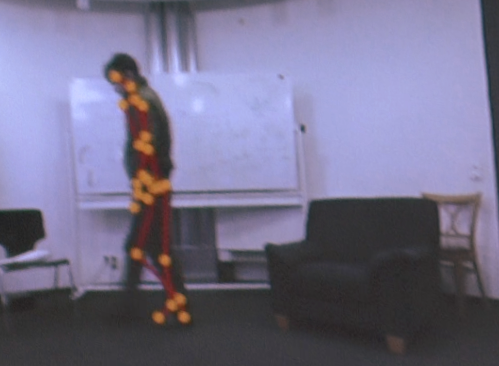}
\includegraphics[width=0.192\columnwidth,trim=1cm 0cm 4.5cm 0cm,clip]{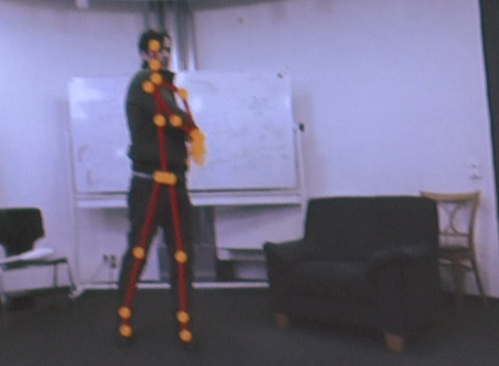}
\caption{Side-by-side pose comparison with our method (top) and Kinect (bottom). 
Overall estimated poses are of similar quality (first two frames). Both the Kinect (third and fourth frames) and our approach (fourth and fifth frames) occasionally predict erroneous poses.
}
\label{fig:livingroom_simple}
\end{figure}

We synchronously recorded video from an RGB camera and a co-located Kinect sensor in a living room scenario.
Figure \ref{fig:livingroom_simple} shows representative frames.
Although the depth sensor provides additional information, our reconstructions from just RGB are of a similar quality.
The Kinect results are of comparable stability to ours, but yield erroneous reconstructions when limbs are close to scene objects, such as when sitting down.
Our RGB solution, however, succeeds in this case, although is slightly less reliable in depth estimation.
A challenging case for both methods is the tight crossing of legs.
\change{Please see the supplemental video for a visual comparison.}
%% A live demonstration is shown in the video.

\begin{figure}[]
\center
\includegraphics[width=\linewidth]{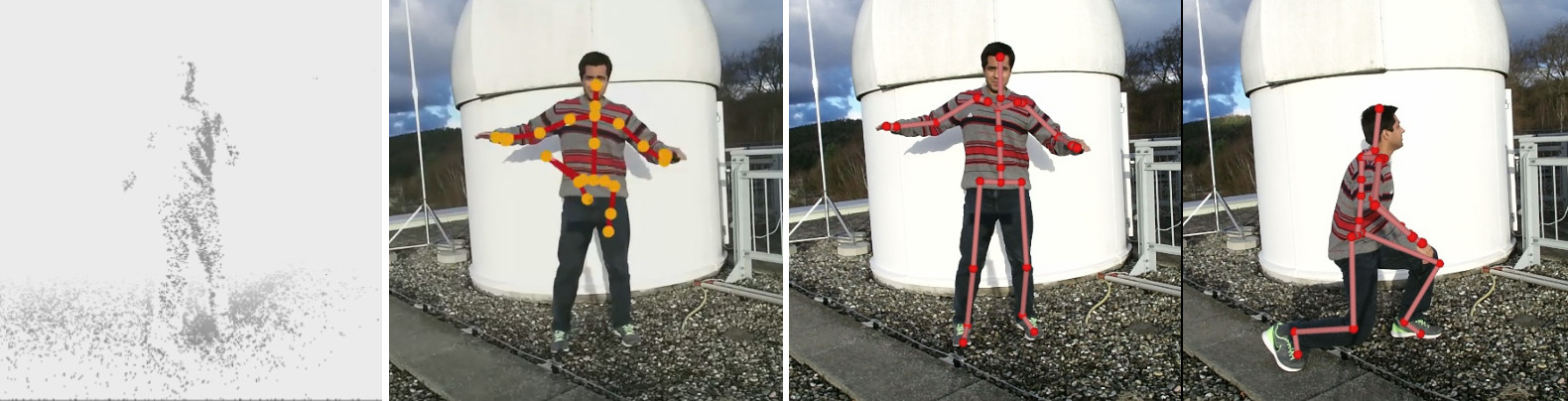}
\caption{Our approach succeeds in strong illumination and sunlight (center right and right), while the IR-based depth estimates of the Microsoft Kinect are erroneous (left) and depth-based tracking fails (center left).}
\label{fig:livingroom_illuminated}
\end{figure}

The video solution succeeds also in situations with direct sunlight (Figure~\ref{fig:livingroom_illuminated}), where IR-based depth cameras are inoperable.
Moreover, RGB cameras can simply be equipped with large field-of-view (FOV) lenses and, despite strong distortions, successfully track humans~\cite{rhodin_egocap_SIGGRAPHAsia2016}.
% .
On the other hand, existing active sensors are limited to relatively small FOVs, which severely limits the tracking volume.

\subsection{Comparison with Video Solutions}
\begin{figure}
\center
\includegraphics[width=\columnwidth]{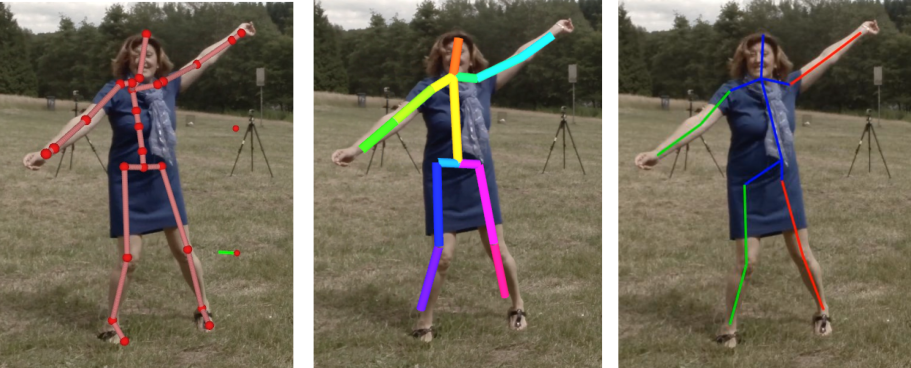}
\caption{Side-by-side comparison of our full method (left), against the offline joint-position estimation methods of Mehta \etal \shortcite{mehta_mlc3d_arxiv16} (middle) and Zhou \etal \shortcite{zhou_convexrelaxation_cvpr2015} (right). Our real-time results are of a comparable quality to these offline methods. 2D joint positions for Zhou \etal are generated using Convolutional Pose Machines \shortcite{wei_cpm_cvpr16}.}
\label{fig:ours_mehta_zhou}
\end{figure}

\parahead{Qualitative Evaluation}
We qualitatively compare against the 3D pose regression method of Mehta~\etal~\shortcite{mehta_mlc3d_arxiv16} and Zhou~\etal~\shortcite{zhou_convexrelaxation_cvpr2015}
on Sequence 6 (outdoor) of MPI-INF-3DHP test set. 
Our results are comparable to the quality of these offline methods (see Figure \ref{fig:ours_mehta_zhou}).
However, the per frame estimates of these offline methods exhibits jitter over time, a drawback of most existing solutions.
Our full pose results are temporally stable and are computed at real-time frame rate of \FPS~Hz.

\begin{figure}[]
\center
\includegraphics[width=\linewidth]{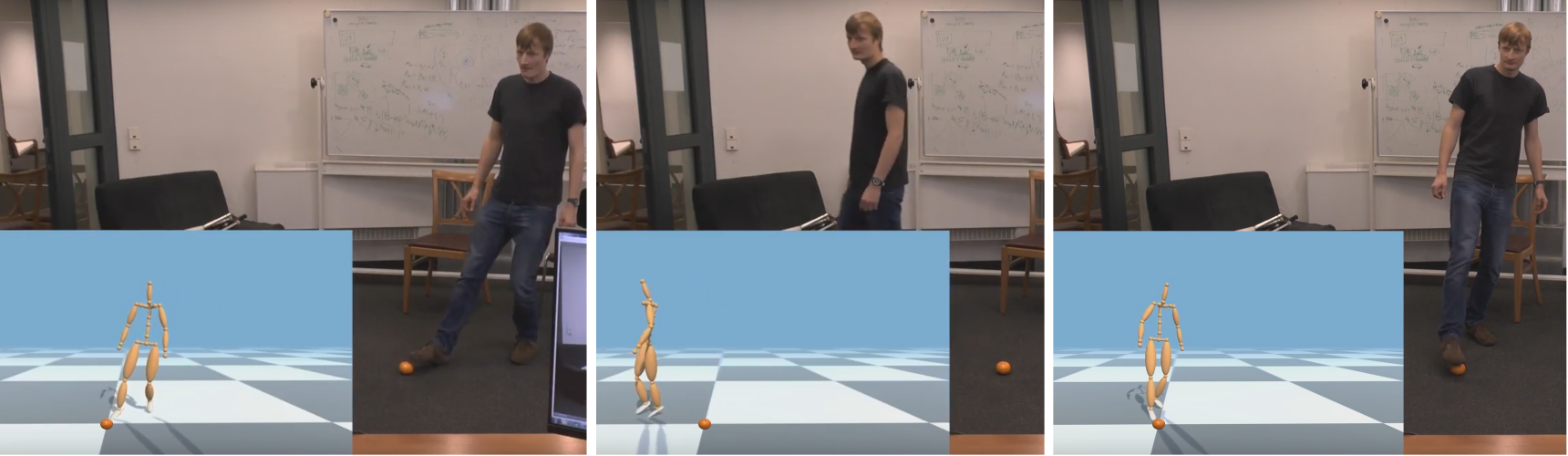}
\caption{The estimated 3D pose is drift-free. The motion of the person starts and ends at the marked point (orange), both in the real world and in our reconstruction.}
\label{fig:driftFree}
\end{figure}

%% \parahead{Global pose}
The kinematic skeleton fitting estimates global translation $\mathbf{d}$.
Figure \ref{fig:driftFree} demonstrates that estimates are drift-free, the feet position matches with the same reference point after performing a circular walk.
The smoothness constraint in depth direction limits sliding of the character away from the character, as pictured in the supplemental video sequences.

\begin{table*}[]
\centering
\caption{Comparison of our network against state of the art on MPI-INF-3DHP test set, using ground-truth bounding boxes. We report the Percentage of Correct Keypoints measure in 3D, and the Area Under the Curve for the same, as proposed by MPI-INF-3DHP. We additionally report the Mean Per Joint Position Error in mm. Higher PCK and AUC is better, and lower MPJPE is better.}
\label{tbl:our_testset}
\begin{tabular}{c|c||c|c|c|c|c|c|c||ccc}
                                                                          &          & Stand/        &               & Sit On        & Crouch/       & On the        &               &               &                           & \multicolumn{1}{l}{}      & \multicolumn{1}{l}{} \\
Network                                                                   & Scales   & Walk          & Exercise      & Chair         & Reach         & Floor         & Sports        & Misc.         & \multicolumn{3}{c}{Total}                                                    \\ \hline
\T                                                                          &          & PCK           & PCK           & PCK           & PCK           & PCK           & PCK           & PCK           & \multicolumn{1}{c|}{PCK}  & \multicolumn{1}{c|}{AUC}  & MPJPE(mm)                \\ \hline \cline{3-12}
\multirow{2}{*}{\begin{tabular}[c]{@{}c@{}}Ours \\ (ResNet 100)\end{tabular}} & 0.7, 1.0 \T & {87.6} & 76.4          & 71.4       & 71.6          & 47.8          & {82.5} & {78.9} & \multicolumn{1}{c|}{75.0} & \multicolumn{1}{c|}{39.5} & 127.8                \\ \cline{2-12}
\T                                                                          & 1.0      & 86.4          & 72.3         & 68.0          & 65.4          & 40.7          & 80.5          & 76.3          & \multicolumn{1}{c|}{71.4} & \multicolumn{1}{c|}{36.9} & 142.8                \\ \hline
\multirow{2}{*}{\begin{tabular}[c]{@{}c@{}}Ours\\ (ResNet 50)\end{tabular}}   & 0.7, 1.0 \T & \textbf{87.7}          & \textbf{77.4}          & 74.7          & 72.9          & 51.3          & \textbf{83.3}          & \textbf{80.1}          & \multicolumn{1}{c|}{\textbf{76.6}} & \multicolumn{1}{c|}{\textbf{40.4}} & 124.7                \\ \cline{2-12}
\T                                                                          & 1.0      & 86.7          & 73.9          & 69.8          & 66.1          & 44.7          & 82.0          & 79.4          & \multicolumn{1}{c|}{73.3} & \multicolumn{1}{c|}{37.8} & 138.7                \\ \hline \hline
\multirow{2}{*}{\begin{tabular}[c]{@{}c@{}}\cite{mehta_mlc3d_arxiv16}\\ (ResNet 100)\end{tabular}}                                              & 0.7, 1.0 \T & 86.6          & {75.3} & \textbf{74.8} & \textbf{73.7} & \textbf{52.2} & 82.1          & 77.5          & \multicolumn{1}{c|}{75.7} & \multicolumn{1}{c|}{39.3} & 117.6                \\\cline{2-12}
\T                                                                          & 1.0      & 86.3          & 72.4          & 71.5          & 67.6          & 49.2          & 81.0          & 76.2          & \multicolumn{1}{c|}{73.2} & \multicolumn{1}{c|}{37.8} & 126.6               
\end{tabular}
\end{table*}

\parahead{Quantitative Evaluation}
\label{sec:quantitative}
We compare our method with the state-of-the-art approach of Mehta~\etal~\shortcite{mehta_mlc3d_arxiv16} on the MPI-INF-3DHP dataset, using the more robust Percentage of Correct Keypoints metric (3D PCK @150mm) on the 14 joints spanned by head, neck, shoulders, elbow, wrist, hips, knees and ankles. We train both, our model, as well as that of Mehta \etal on the same data (Human3.6m + MPI-INF-3DHP), as detailed in Section \ref{sec:network}, to be compatible in terms of the camera viewpoints selected, and use ResNet100 as the base architecture for a fair comparison. Table \ref{tbl:our_testset} shows the results of the raw 3D predictions from our network on ground-truth bounding box cropped frames. We see that the results are comparable to that of Mehta et al. \change{The slight increase in accuracy on going to a 50-layer network is possibly due to the better gradient estimates coming from larger mini-batches that can be fit into memory while training, on account of the smaller size of the network.} Evidence that our method ties the estimated 3D positions strongly to image appearance than previous methods can also be gleaned from the fact that our approach performs significantly better for activity classes such as Standing/Walking, Sports and Miscellaneous without significant self-occlusions. We do lose some performance on activity classes with significant self-occlusion such as Sitting/Lying on the floor. We additionally report the Mean Per Joint Position Error (MPJPE) numbers in mm. Note that MPJPE is not a robust measure, and is heavily influenced by large outliers, and hence the \change{worse performance on} the MPJPE measure (124.7mm vs 117.6mm) \change{despite the better 3D PCK results (76.6\% vs 75.7\%).} 

\begin{figure}
  %% \centering\includegraphics*[page=1,width=1.0\linewidth]{Figures/comparison.png}
    \centering\includegraphics*[page=1,width=0.9\linewidth]{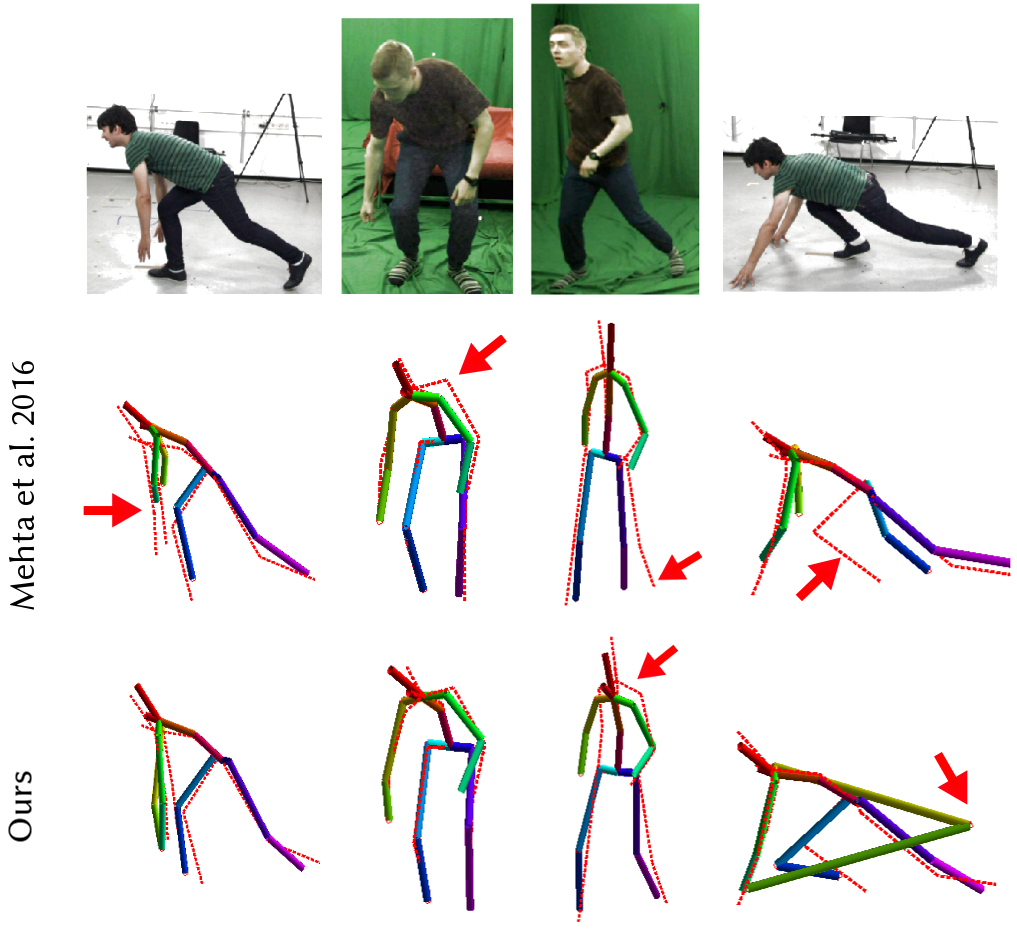}
  \caption{A visual look at the direct 3D predictions resulting from our fully-convolutional formulation vs Mehta \etal
          Our formulation allows the predictions to be more strongly tied to image evidence, leading to overall better pose quality, particular for the end effectors.  The red arrows point to mispredictions.}
  \label{fig:comparison}
\end{figure}
\begin{figure}[t!]
  	\centering\includegraphics*[page=1,width=0.95\linewidth]{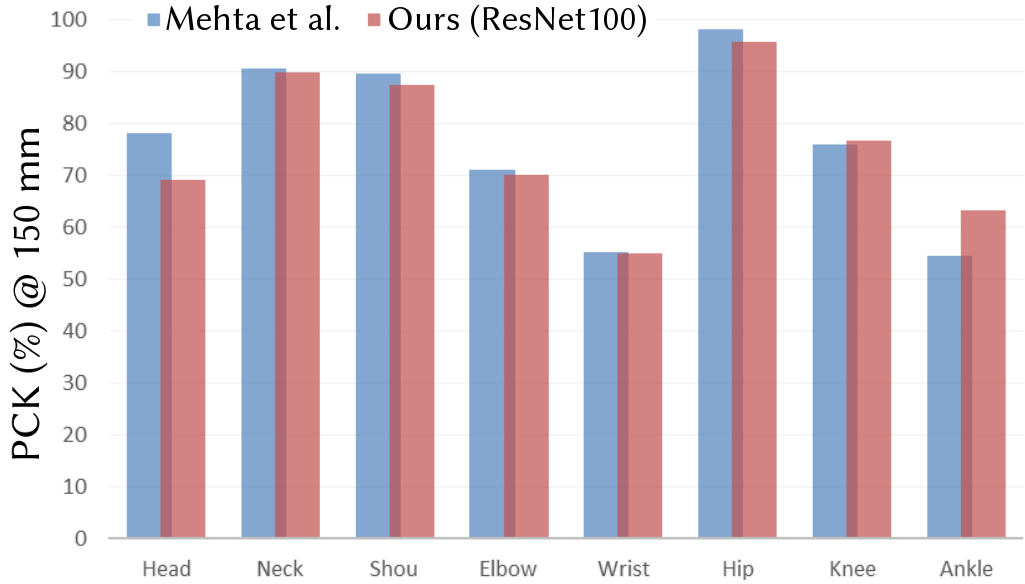}
	\caption{Joint-wise breakdown of the accuracy of Mehta \etal and Our ResNet100 based CNN predictions on MPI-INF-3DHP test set.}
	\label{fig:comparison_chart}
\end{figure}
We further investigate the nature of errors of our method. We first look at the joint-wise breakup of accuracy of our fully-convolutional ResNet100 CNN predictions vs Mehta \etal's formulation with fully-connected layers. Figure \ref{fig:comparison_chart} shows that the accuracy of ankles for our formulation is significantly better, while the accuracy of the head is markedly worse. 

\begin{figure}[]
   \centering\includegraphics*[page=1,width=1.0\linewidth]{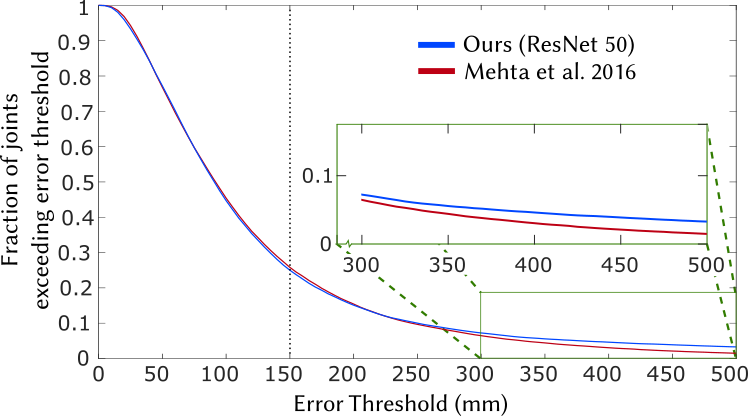}
  \caption{Fraction of joints incorrectly predicted on MPI-INF-3DHP test set, as determined by the distance between the predicted joint location and the ground truth joint location being greater than the error threshold. The dotted line marks the threshold for which the 3D PCK numbers are reported. At bottom right we see that our method has larger occasional mispredictions, which result in higher MPJPE numbers despite otherwise similar performance.}
  \label{fig:auc_curve}
\end{figure}

In Figure \ref{fig:comparison}, we visually compare the two methods, further demonstrating the strong tie-in to image appearance that our formulation affords, and the downsides of the strong tie-in. We also show that our method is prone to occasional large mispredictions when the body joint 2D location detector misfires. It is these large outliers that obfuscate the reported MPJPE numbers. \change{Figure \ref{fig:auc_curve}, which plots the fraction of mispredicted joints vs. the error threshold on MPI-INF-3DHP test set shows that our method has a higher fraction of per-joint mispredictions beyond 300mm. It explains the higher MPJPE numbers compared to Mehta \etal despite equivalent PCK performance.} The various filtering stages employed in the full pipeline ameliorate these large mispredictions.   

For Human3.6m, we follow the protocol as in earlier work \cite{pavlakos_volumetric3d_arxiv16,tekin_fusion_arxiv16,tekin_motion_comp_cvpr16}, and evaluate on all actions and cameras for subject number 9 and 11, and report Mean Per Joint Position Error (mm) for root relative 3D joint positions from our network. See Table \ref{tbl:h34m}. Note that despite the occasional large outliers affecting the MPJPE measure, our predictions are still better than most of the existing methods.

The accuracy attained from single view methods is still below that of real-time multi-view methods, which can achieve a mean accuracy of the order of 10mm \cite{stoll_fast_iccv2011}.

\begin{table}
\centering
\caption{Results on MPI-INF-3DHP test set with the bounding box corners randomly jittered between +/- 40px to emulate noise from a BB estimator. Our fully-convolutional formulation is more robust than a comparative fully-connected formulation. The evaluation is at a single scale (1.0).}
\label{tbl:jitter}
\renewcommand{\tabcolsep}{1pt}
\resizebox{1.01\columnwidth}{!}{
\begin{tabular}{l|c|c|c|c|c|c|c||cc}
                            & \multicolumn{1}{c|}{Stand/}               &    \multicolumn{1}{c|}{ }                 & \multicolumn{1}{c|}{Sit On}               & \multicolumn{1}{c|}{Crouch/}              & \multicolumn{1}{c|}{On the}               &   \multicolumn{1}{c|}{ }               & \multicolumn{1}{c||}{}     & \multicolumn{1}{|l}{}      & \multicolumn{1}{l}{} \\
\multicolumn{1}{c|}{Network} & \multicolumn{1}{c|}{Walk}                 & \multicolumn{1}{c|}{Exerc.}             & \multicolumn{1}{c|}{Chair}                & \multicolumn{1}{c|}{Reach}                & \multicolumn{1}{c|}{Floor}               & \multicolumn{1}{c|}{Sport }              & \multicolumn{1}{c||}{Misc.} & \multicolumn{2}{|c}{Total}                        \\ \hline
         \T                   & \multicolumn{1}{c|}{PCK}                  & \multicolumn{1}{c|}{PCK}                  & \multicolumn{1}{c|}{PCK}                  & \multicolumn{1}{c|}{PCK}                  & \multicolumn{1}{c|}{PCK}                  & \multicolumn{1}{c|}{PCK}                  & \multicolumn{1}{c||}{PCK}  & \multicolumn{1}{c|}{~PCK~}  & ~AUC~                  \\ \hline
\multicolumn{1}{l|}{Ours (ResNet 100)\T} & 86.0                 & 71.0                 & 65.0                 & 61.1                 & 37.4                 & 78.9                 & \multicolumn{1}{c||}{75.5} & \multicolumn{1}{c|}{69.5} & 35.8                 \\
\multicolumn{1}{l|}{Ours (ResNet 50)}  & 84.9                 & 69.4                 & 65.1                 & 61.9                 & 40.8                 & 78.6                 & \multicolumn{1}{c||}{77.6} & \multicolumn{1}{c|}{70.1} & 35.7               \\ \hline
\cite{mehta_mlc3d_arxiv16}\T    & 81.2                 & 64.2                 & 67.1                 & 62.1                 & 43.5                 & 76.0                 & \multicolumn{1}{c||}{71.1} & \multicolumn{1}{c|}{67.8} & 34.0                \\\hline
\end{tabular}
}
\end{table}

\begin{table*}[]
\centering
\caption{Results of our raw CNN predictions on Human3.6m, evaluated on the ground truth bounding box crops for all frames of Subject 9 and 11. Our CNNs use only Human3.6m as the 3D training set, and are pretrained for 2D pose prediction. The error measure used is Mean Per Joint Position Error (MPJPE) in millimeters. Note again that the error measure used is not robust, and subject to obfuscation from occasional large mispredictions, such as those exhibited by our raw CNN predictions.}
\label{tbl:h34m}
\renewcommand{\tabcolsep}{2pt}
\begin{tabular}{l||c|c|c|c|c|c|c|c|c|c|c|c|c|c|c||c}
                &        &         &        &       &       &        &        &         & Sit   &       & Take  &       &       & Walk  & Walk  & \multicolumn{1}{l}{} \\
\multicolumn{1}{c||}{Method}                      & Direct & Discuss & Eating & Greet & Phone & Posing & Purch. & Sitting & Down  & Smoke & Photo & Wait  & Walk  & Dog   & Pair  & All                  \\ \hline \hline
\cite{zhou_sparseness_deepness_cvpr15}     & 87.4   & 109.3   & 87.1   & 103.2 & 116.2 & 106.9  & 99.8   & 124.5   & 199.2 & 107.4 & 143.3 & 118.1 & 79.4  & 114.2 & 97.7  & 113.0                \\
\cite{tekin_motion_comp_cvpr16}   & 102.4  & 147.7   & 88.8   & 125.3 & 118.0 & 112.3  & 129.2  & 138.9   & 224.9 & 118.4 & 182.7 & 138.8 & 55.1  & 126.3 & 65.8  & 125.0                \\
\cite{yu_mono_heightmap_eccv16}          & 85.1   & 112.7   & 104.9  & 122.1 & 139.1 & 105.9  & 166.2  & 117.5   & 226.9 & 120.0 & 135.9 & 117.7 & 137.4 & 99.3  & 106.5 & 126.5                \\ \hline
\cite{ionescu_human36_pami14}                & 132.7  & 183.6   & 132.4  & 164.4 & 162.1 & 150.6  & 171.3  & 151.6   & 243.0 & 162.1 & 205.9 & 170.7 & 96.6  & 177.1 & 127.9 & 162.1                \\
\cite{zhou_deep_kinematic_arxiv16}       & 91.8   & 102.4   & 97.0   & 98.8  & 113.4 & 90.0   & 93.8   & 132.2   & 159.0 & 106.9 & 125.2 & 94.4  & 79.0  & 126.0 & 99.0  & 107.3                \\
\cite{pavlakos_volumetric3d_arxiv16} & 58.6   & 64.6    & 63.7   & 62.4  & 66.9  & 57.7   & \textbf{62.5}   & \textbf{76.8}    & \textbf{103.5} & \textbf{65.7}  & \textbf{70.7}  & \textbf{61.6}  & 69.0  & \textbf{56.4}  & 59.5  & \textbf{66.9}                 \\
\cite{mehta_mlc3d_arxiv16}          & \textbf{52.6}   & \textbf{63.8}    & \textbf{55.4}   & \textbf{62.3}  & 71.8  & \textbf{52.6}   & 72.2   & 86.2    & 120.6 & 66.0  & 79.8  & 64.0  & \textbf{48.9}  & 76.8  & \textbf{53.7}  & 68.6                 \\
\cite{tekin_fusion_arxiv16}        & 85.0   & 108.8   & 84.4   & 98.9  & 119.4 & 98.5   & 93.8   & 73.8    & 170.4 & 85.1  & 95.7  & 116.9 & 62.1  & 113.7 & 94.8  & 100.1                \\ \hline \hline
Ours (ResNet 100)   & 61.7   & 77.8    & 64.6   & 70.3  & 90.5  & 61.9   & 79.8   & 113.2   & 153.1 & 80.9  & 94.4  & 75.1  & 54.9  & 83.5  & 61.0  & 82.5                 \\
Ours (ResNet 50)   & 62.6   & 78.1    & 63.4   & 72.5  & 88.3  & 63.1   & 74.8   & 106.6   & 138.7 & 78.8  & 93.8  & 73.9  & 55.8  & 82.0  & 59.6  & 80.5                
\end{tabular}
\end{table*}

\begin{figure}[]
   \centering\includegraphics*[page=1,width=1.0\linewidth]{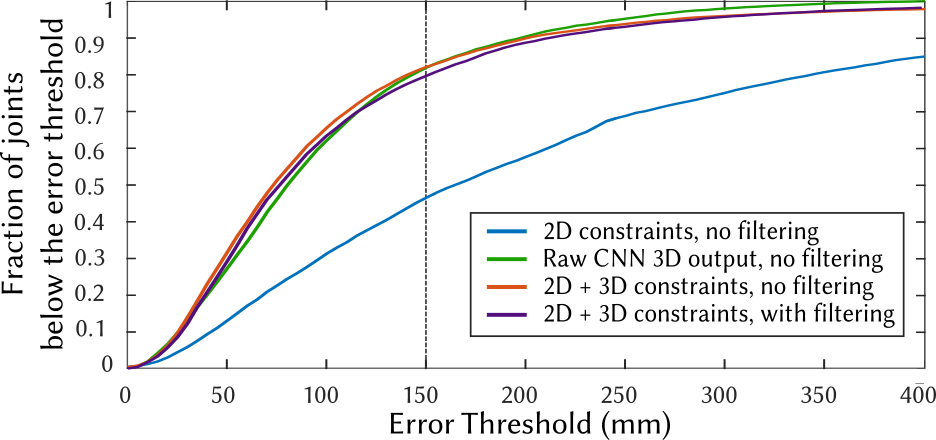}
	\caption{Fraction of joints correctly predicted on the TS1 sequence of MPI-INF-3DHP test set, as determined by the distance between the predicted joint location and the ground truth joint location being below the error threshold. The dotted line marks the 150mm threshold for which the 3D PCK numbers are reported. We see that only using the 2D predictions as constraints for skeleton fitting (blue) performs significantly worse than using both 2D and 3D predictions as constraints (red). Though adding 1 Euro filtering (purple) visually improves the results, the slightly higher error here is due to the sluggish recovery from tracking failures. The 3D predictions from the CNN (green) are also shown.}
    \vspace{-0.4cm}
	\label{fig:ablative_auc}
\end{figure}

\parahead{Generalization to Different Persons and Scenes}
\label{sec:generalization}
We tested our method on a variety of actors, it succeeds for different body shapes, gender and skin tone. See supplemental video. To further validate the robustness we applied the methods to community videos from YouTube, see Figure \ref{fig:teaser}. It generalizes well to the different backgrounds and camera types.

\parahead{Model Components}
\label{sec:ablation}
To demonstrate that our fully-convolutional pose formulation is less sensitive to inexact cropping than networks using a fully-connected formulation, we emulate a noisy BB estimator by jittering the ground-truth bounding box corners of MPI-INF-3DHP test set uniformly at random in the range of +/- 40~px. This also captures scenarios where one or more end effectors are not in the frame, so a loss in accuracy is expected for all methods. Table \ref{tbl:jitter} shows that the fully-connected formulation of Mehta \etal suffers a worse hit in accuracy than our approach, going down by 7.9 PCK, while our comparable network goes down by only 5.5 PCK.

\change{We show the effect of the various components of our full pipeline on the TS1 sequence of MPI-INF-3DHP test set in Figure \ref{fig:ablative_auc}. Without the $E_{\text{IK}}$ component of $E_{\text{total}}$ the tracking accuracy goes down to a PCK of 46.1\% compared to a PCK of 81.7\% when $E_{\text{IK}}$ is used. The raw CNN 3D predictions in conjunction with the BB tracker result in a PCK of 80.3\%. Using $E_{\text{IK}}$ in $E_{\text{total}}$ produces consistently better results for all thresholds lower than 150~mm. This shows the improvements brought about by our skeleton fitting term. 
Additionally, as shown in the supplementary video, using 1 Euro filtering produces qualitatively better results, but the overall PCK decreases slightly (79.7\%) due to slower recovery from tracking failures. 
}
The influence of the smoothness and filtering steps on the temporal consistency are further analyzed in the supplemental video.

\subsection{Applications}
\label{sec:applications}
Our approach is suitable for various interactive applications since it is real-time, temporally stable, fully automatic, and exports data directly in a format amenable to 3D character control.

\begin{figure}
\center
\includegraphics[width=0.23\columnwidth]{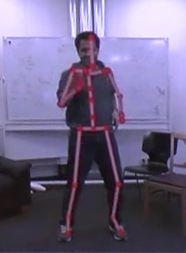}
\includegraphics[width=0.23\columnwidth]{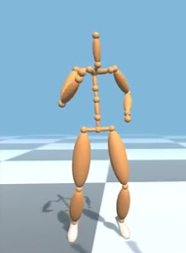}
\hspace{0.15cm}
\includegraphics[width=0.23\columnwidth]{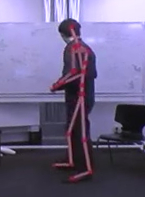}
\includegraphics[width=0.23\columnwidth]{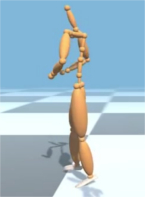}

\vspace{0.25cm}

\includegraphics[width=0.23\columnwidth]{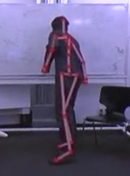}
\includegraphics[width=0.23\columnwidth]{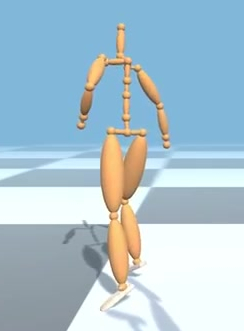}
\hspace{0.15cm}
\includegraphics[width=0.23\columnwidth]{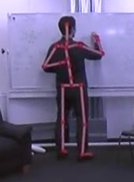}
\includegraphics[width=0.23\columnwidth]{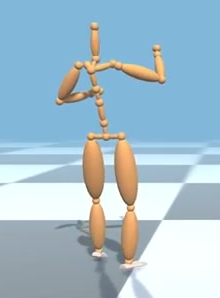}

\vspace{0.25cm}

\includegraphics[width=0.23\columnwidth]{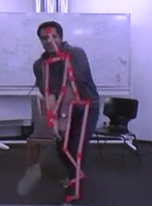}
\includegraphics[width=0.23\columnwidth]{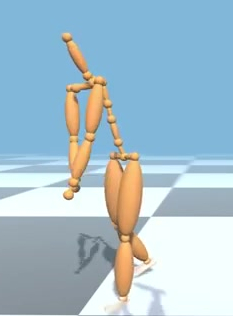}
\hspace{0.15cm}
\includegraphics[width=0.23\columnwidth]{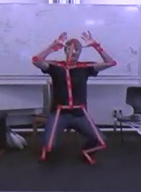}
\includegraphics[width=0.23\columnwidth]{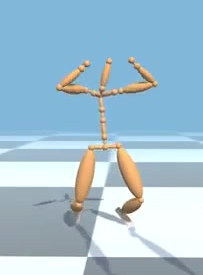}

\vspace{0.25cm}

\includegraphics[width=0.23\columnwidth]{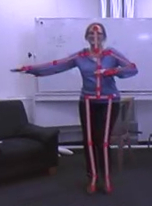}
\includegraphics[width=0.23\columnwidth]{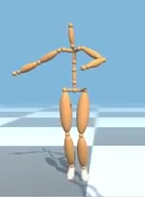}
\hspace{0.15cm}
\includegraphics[width=0.23\columnwidth]{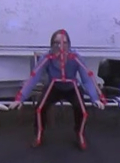}
\includegraphics[width=0.23\columnwidth]{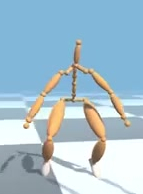}
%\includegraphics[width=\columnwidth]{Figures/helge-boxing01.png}
%\framebox[\columnwidth]{FIGURE: entertainment: golf swing and virtual avatar}
\caption{Application to entertainment. The real-time 3D pose estimation method provides a natural motion interface, e.g.~for sport games.}
\label{fig:entertainment}
\end{figure}

\parahead{Character Control}
Real-time motion capture solutions provide a natural interface
\change{for} game characters and virtual avatars, which go beyond classical mouse and gamepad control.
We applied our method on \change{motions common in activities like} tennis, dance, and juggling, see Figures \ref{fig:teaser} and \ref{fig:entertainment}.
The swing of the arm and leg motion is nicely captured and could, for instance, be used in a casual sports and dancing game, but also for motion analysis of professional athletes to optimize their motion patterns.
We also show successful results in non front-facing motions such as turning and writing on a wall, as well as squatting.

\parahead{Virtual Reality}
The recent availability of cheap head-mounted displays has sparked a range of new applications.
Many products use handheld devices to track the user's hand position for interaction.
Our solution enables them from a single consumer color camera.
Beyond interaction, our marker-less full-body solution enables embodied virtual reality, see Figure \ref{fig:teaser}.
A rich immersive feeling is created by posing a virtual avatar of the user exactly to \change{their} own real pose.
With our solution the real and virtual pose are aligned such that users perceive the virtual body as their own.

\begin{figure}[]
\centering
\includegraphics[width=\columnwidth]{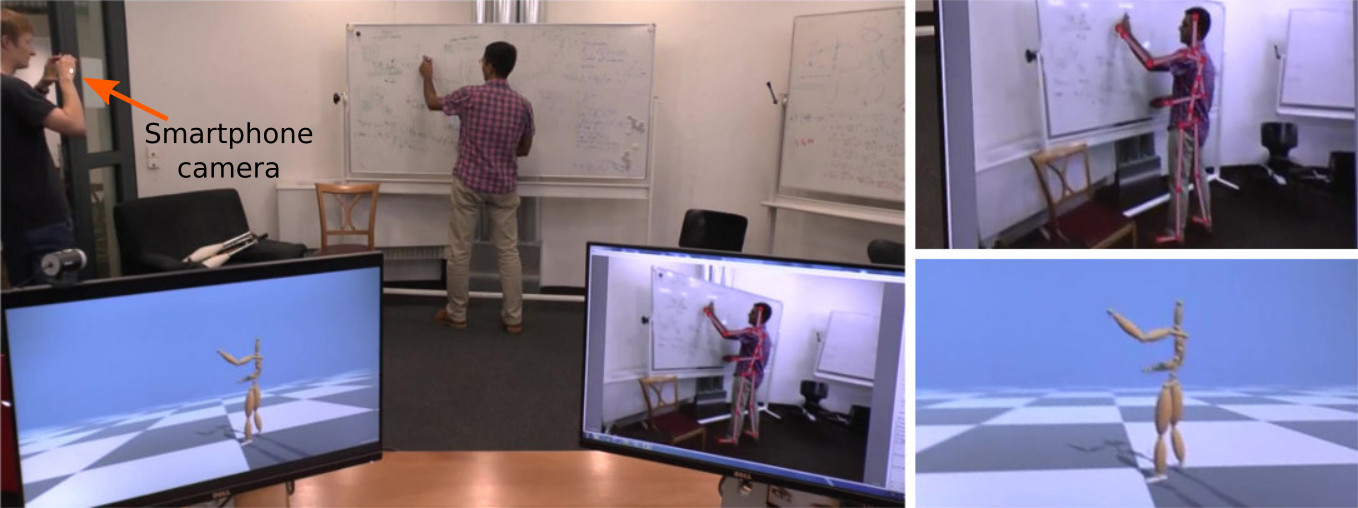}
\vspace{0.2cm}
\includegraphics[width=\columnwidth]{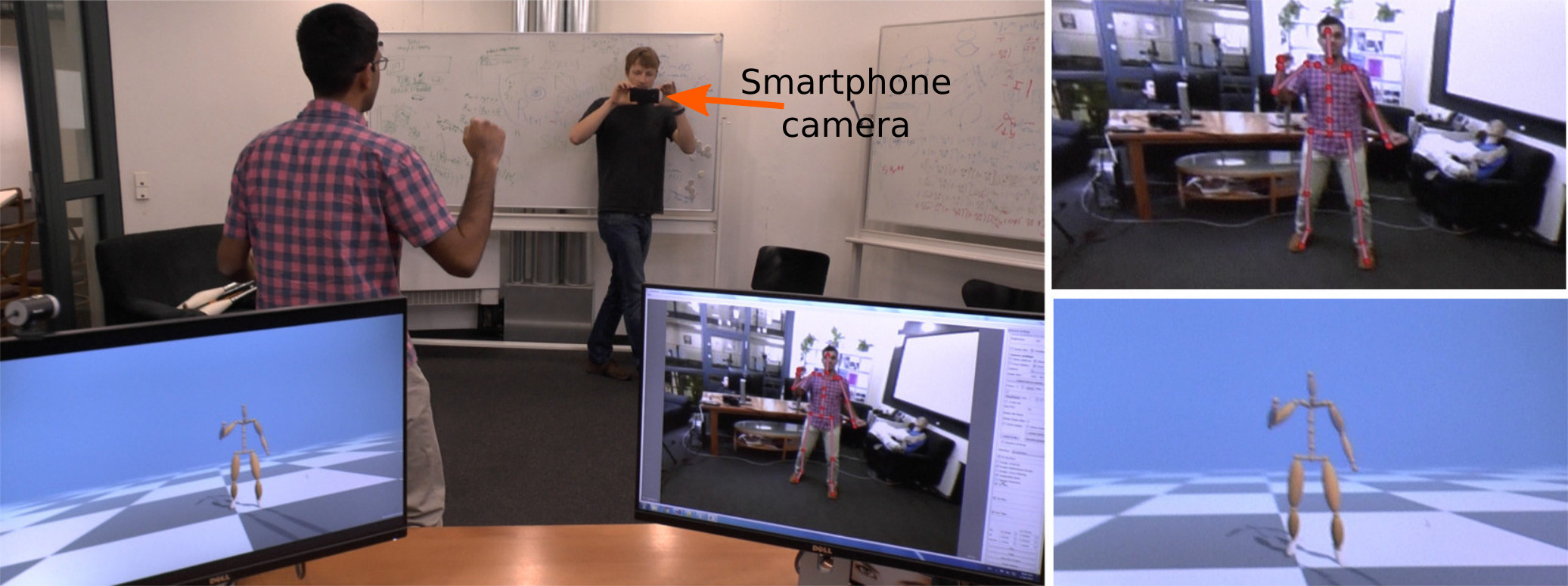}
\caption{Handheld recording with a readily available smartphone camera (left) and our estimated pose (right), streamed to and processed by a GPU enabled PC.}
\vspace{-0.5cm}

\label{fig:smartphone-capture}
\end{figure}

\parahead{\change{Ubiquitous Motion Capture with Smartphones}}
Real-time monocular 3D pose estimation lends itself to application on \change{low quality} smartphone video streams.
By streaming the video to a machine with sufficient capabilities for our algorithm, one can turn any smartphone into a lightweight, fully-automatic, handheld motion capture sensor, see Figure~\ref{fig:smartphone-capture} and the accompanying video.
Since smartphones are widespread, it enables the aforementioned applications for casual users without requiring additional sensing devices.

\section{Limitations}

Depth estimation from a monocular image is severely ill posed, slight inaccuracies in the estimation can lead to largely different depth estimates, which manifests also in our results in slight temporal jitter.
We claim improved stability and temporal consistency compared to existing monocular RGB 3D pose estimation methods.
This uncertainty could be further reduced with domain specific knowledge, e.g., foot-contact constraints when the floor location is known, and head-pose stabilization with the position of head-mounted-displays in VR applications, which is readily obtained with IMU-sensors. 

A downside of our CNN prediction formulation is that mispredictions of 2D joint locations result in implausible 3D poses. This is ameliorated in the tracker through skeleton retargeting and pose filtering. This could be addressed directly in the CNN through imposition of stronger interdependencies between predictions. Additionally, the performance on poses with significant amounts of self occlusion remains a challenge.  

Further, very fast motions can exceed the convergence radius of our IK optimization, but the integration of per frame 2D and 3D pose yields quick recovery from erroneous poses. \change{Initial experiments with $256\times256$~px input to the CNN show that much higher frame rates are possible with no loss in accuracy.}

\section{Discussion}   
The availability of sufficient annotated 3D pose training data \change{remains an issue. Even the most recent} annotated real 3D pose data sets, or combined real/synthetic data sets~\cite{chen_synth_data_3dv16,ionescu_human36_pami14,mehta_mlc3d_arxiv16} are a subset of real world human pose, shape, appearance and background distributions.
Recent top performing methods explicitly address this data sparsity by training similarly deep networks, but with architectural changes enabling improved intermediate training supervision \cite{mehta_mlc3d_arxiv16}.

Our implementation only supports a single person, although the proposed fully-convolutional formulation could be scaled to multiple persons. Such an extension is currently precluded due to the lack of multi-person datasets, required to train multi-person 3D pose regressors. One possible approach is to adapt the multi-person 2D pose methods of Insafutdinov~\etal~\shortcite{insafutdinov_deepercut_eccv16} and Cao~\etal~\shortcite{cao2016realtime}.

We also analyze the impact of 2D joint position mispredictions on the 3D joint position predictions from our fully-convolutional formulation. We decouple the 3D predictions from the 2D predictions by looking up the 3D joint positions from their location-maps using the ground truth 2D joint positions. See Table~\ref{tbl:gt_2d}. We see a 3D PCK improvement of 2.8, which is congruent with the notion of a stronger tie-in of the predicted joint positions with the image plane, which causes the 3D joint predictions to be erroneous when 2D joint detection misfires. The upside of this is that the 3D predictions can be improved through improvements to 2D joint position prediction. Alternatively, optimization formulations that directly operate on the heatmaps and the location-maps could be constructed.
Our fully-convolutional formulation can also benefit from iterative refinement, akin to heatmap-based 2D pose estimation approaches \cite{newell_stacked_hourglass_eccv16,hu_bottomup_cvpr16}.

\begin{table}[]
\centering
\caption{Results on MPI-INF-3DHP test set with the 3D joint position lookup in the location-maps done using the ground truth 2D locations rather the predicted 2D locations. We see that the location maps have captured better 3D pose information, which can perhaps be extracted through optimization methods operating directly on heatmaps and location-maps. The evaluation uses 2 scales (0.7, 1.0).
}
\label{tbl:gt_2d}
\renewcommand{\tabcolsep}{1pt}
\resizebox{1.01\columnwidth}{!}{
\begin{tabular}{l|c|c|c|c|c|c|c||cc}
                            & \multicolumn{1}{c|}{Stand/}               &    \multicolumn{1}{c|}{ }                 & \multicolumn{1}{c|}{Sit On}               & \multicolumn{1}{c|}{Crouch/}              & \multicolumn{1}{c|}{On the}               &   \multicolumn{1}{c|}{ }               & \multicolumn{1}{c||}{}     & \multicolumn{1}{|l}{}      & \multicolumn{1}{l}{} \\
\multicolumn{1}{c|}{Network} & \multicolumn{1}{c|}{Walk}                 & \multicolumn{1}{c|}{Exerc.}             & \multicolumn{1}{c|}{Chair}                & \multicolumn{1}{c|}{Reach}                & \multicolumn{1}{c|}{Floor}               & \multicolumn{1}{c|}{Sport }              & \multicolumn{1}{c||}{Misc.} & \multicolumn{2}{|c}{Total}                        \\ \hline
         \T                   & \multicolumn{1}{c|}{PCK}                  & \multicolumn{1}{c|}{PCK}                  & \multicolumn{1}{c|}{PCK}                  & \multicolumn{1}{c|}{PCK}                  & \multicolumn{1}{c|}{PCK}                  & \multicolumn{1}{c|}{PCK}                  & \multicolumn{1}{c||}{PCK}  & \multicolumn{1}{c|}{~PCK~}  & ~AUC~                  \\ \hline
\multicolumn{1}{l|}{Ours (ResNet 100)\T} &88.1 & 80.9 & 74.0 & 76.1 & 56.3 & 82.9 & \multicolumn{1}{c||}{80.2} & \multicolumn{1}{c|}{77.8} & 41.0                 \\
\multicolumn{1}{l|}{Ours (ResNet 50)}  & 88.0 & 81.8 & 78.6 & 77.4 & 59.3 & 82.8 & \multicolumn{1}{c||}{81.2} & \multicolumn{1}{c|}{79.4} & 41.6              \\ \hline \hline
\multicolumn{1}{l|}{\cite{mehta_mlc3d_arxiv16}\T}  & 86.6 & 75.3 & 74.8 & 73.7 & 52.2 & 82.1 & \multicolumn{1}{c||}{77.5} & \multicolumn{1}{c|}{75.7} & 39.3    \\ \hline
\end{tabular}
}
\end{table}

\section{Conclusion}
We have presented the first method that estimates the 3D \change{kinematic} pose of a human, including global position, \change{in a stable, temporally consistent manner} from a single RGB video stream at 30~Hz.
Our approach combines a fully-convolutional CNN that regresses 2D and 3D joint positions and a kinematic skeleton fitting method, producing a real-time temporally stable 3D reconstruction of the motion.
In contrast to most existing approaches, our network can operate without strict bounding boxes, and facilitates inexpensive bounding box tracking.
We have shown results in a variety of challenging real-time scenarios, including live streaming from a smartphone camera, as well as in community videos. 
A number of applications have been presented, such as embodied VR and interactive character control for computer games.

Qualitative and quantitative evaluations demonstrate that our approach compares to offline state-of-the-art monocular RGB methods and approaches the quality of real-time \mbox{RGB-D} methods. 
Hence, we believe our method is a significant step forward to democratizing 3D human pose estimation, lifting both the need for special cameras such as the IR-based depth cameras, as well as the long and heavy processing times.

\bibliographystyle{ACM-Reference-Format}
%\nocite{*}
\bibliography{article}
\end{document}